\let\TeXyear\year
\documentclass{ieeeaccess}

\let\year\TeXyear

\usepackage{cite}
\usepackage{amsmath,amssymb,amsfonts}
\usepackage{algorithmic}
\usepackage{graphicx}
\usepackage{textcomp}

\newcommand{\qemph}[1]{\emph{#1}}

\usepackage{cite}
\usepackage{textcomp}

\usepackage{pgfplots}
\usepackage{pgfplotstable}
\def\BibTeX{{\rm B\kern-.05em{\sc i\kern-.025em b}\kern-.08em
    T\kern-.1667em\lower.7ex\hbox{E}\kern-.125emX}}

\usepackage{xcolor}
\usepackage[T1]{fontenc} 
\usepackage[utf8]{inputenc} 
\usepackage[english]{babel} 

\usepackage{amsmath} 
\usepackage{amsfonts}
\usepackage{amssymb}  

\usepackage{diagbox}
\usepackage{makecell}
\usepackage{multirow}

\usepackage[inline]{enumitem}

\usepackage{fontawesome}
\usepackage{bclogo}

\usepackage{booktabs} 

\usepackage{dsfont}
\usepackage{graphicx}
\usepackage{subcaption} 

\usepackage{amsmath}
\usepackage{pifont}
\usepackage{booktabs}
\usepackage{siunitx}
\usepackage{tikz}
\usetikzlibrary{shapes.geometric, calc}

\newcommand\score[2]{%
  \pgfmathsetmacro\pgfxa{#1 + 1}%
  \tikzstyle{scorestars}=[star, star points=5, star point ratio=2.25, draw, inner sep=0.15em, anchor=outer point 3]%

  \begin{tikzpicture}[lineline]
    \foreach \i in {1, ..., #2} {
      \pgfmathparse{\i<=#1 ? "yellow" : "gray"}
      \edef\starcolor{\pgfmathresult}
      \draw (\i*1em, 0) node[name=star\i, scorestars, fill=\starcolor]  {};
    }
    \pgfmathparse{#1>int(#1) ? int(#1+1) : 0}
    \let\partstar=\pgfmathresult
    \ifnum\partstar>0
      \pgfmathsetmacro\starpart{#1-(int(#1)}
      \path [clip] ($(star\partstar.outer point 3)!(star\partstar.outer point 2)!(star\partstar.outer point 4)$) rectangle 
      ($(star\partstar.outer point 2 |- star\partstar.outer point 1)!\starpart!(star\partstar.outer point 1 -| star\partstar.outer point 5)$);
      \fill (\partstar*1em, 0) node[scorestars, fill=yellow]  {};
    \fi
  \end{tikzpicture}%
}

\usepackage{bbding}
\usepackage{pifont}
\newcommand{\xmark}{\ding{55}}%
\usepackage{multirow}

\usepackage{tikz}
\usepackage{pgfplots}
\pgfplotsset{compat=newest}
\usepgfplotslibrary{fillbetween}

\usepackage{epsfig} 
\usepackage{times} 
\usepackage{siunitx}
\usepackage{calc}
\usepackage[hyphens]{url}
\usepackage{svg}
\renewcommand{\baselinestretch}{0.97}

\usepackage{fontawesome}

\usepackage{tabularx}
\usepackage{multirow}
\usepackage{pifont}

\usepackage{enumitem}

\usepackage{tabularx}
\usepackage{bbm}
\usepackage{dsfont}

\usepackage{scalerel}
\usepackage{tikz}

\NewSpotColorSpace{PANTONE}
  \AddSpotColor{PANTONE} {PANTONE3015C} {PANTONE\SpotSpace 3015\SpotSpace C} {1 0.3 0 0.2}
  \SetPageColorSpace{PANTONE}

\usetikzlibrary{svg.path}

\definecolor{orcidlogocol}{HTML}{A6CE39}
\tikzset{
	orcidlogo/.pic={
		\fill[orcidlogocol] svg{M256,128c0,70.7-57.3,128-128,128C57.3,256,0,198.7,0,128C0,57.3,57.3,0,128,0C198.7,0,256,57.3,256,128z};
		\fill[white] svg{M86.3,186.2H70.9V79.1h15.4v48.4V186.2z}
		svg{M108.9,79.1h41.6c39.6,0,57,28.3,57,53.6c0,27.5-21.5,53.6-56.8,53.6h-41.8V79.1z M124.3,172.4h24.5c34.9,0,42.9-26.5,42.9-39.7c0-21.5-13.7-39.7-43.7-39.7h-23.7V172.4z}
		svg{M88.7,56.8c0,5.5-4.5,10.1-10.1,10.1c-5.6,0-10.1-4.6-10.1-10.1c0-5.6,4.5-10.1,10.1-10.1C84.2,46.7,88.7,51.3,88.7,56.8z};
	}
}

\usepackage{arydshln} 
\newcommand{\dashmidrule}{%
	\noalign{\vskip 0.8ex} 
	\hdashline[2.0pt/2pt] 
	\noalign{\vskip 0.8ex} 
}

\newcommand\orcidicon[1]{\href{https://orcid.org/#1}{\mbox{\scalerel*{
				\begin{tikzpicture}[yscale=-1,transform shape]
					\pic{orcidlogo};
				\end{tikzpicture}
			}{|}}}}

\usepackage{hyperref} 

\usepackage{cleveref}

\usepackage{bm}
\makeatletter
\AtBeginDocument{\DeclareMathVersion{bold}
\SetSymbolFont{operators}{bold}{T1}{times}{b}{n}
\SetSymbolFont{NewLetters}{bold}{T1}{times}{b}{it}
\SetMathAlphabet{\mathrm}{bold}{T1}{times}{b}{n}
\SetMathAlphabet{\mathit}{bold}{T1}{times}{b}{it}
\SetMathAlphabet{\mathbf}{bold}{T1}{times}{b}{n}
\SetMathAlphabet{\mathtt}{bold}{OT1}{pcr}{b}{n}
\SetSymbolFont{symbols}{bold}{OMS}{cmsy}{b}{n}
\renewcommand\boldmath{\@nomath\boldmath\mathversion{bold}}}
\makeatother

\def\BibTeX{{\rm B\kern-.05em{\sc i\kern-.025em b}\kern-.08em
    T\kern-.1667em\lower.7ex\hbox{E}\kern-.125emX}}

\usepackage{hologo}
\usepackage{listings}
\begin{document}

\twocolumn[
\begin{@twocolumnfalse}
	\Huge {IEEE copyright notice} \\ \\
	\large{ \textcopyright\ 2026 IEEE. Personal use of this material is permitted. Permission from IEEE must be obtained for all other uses, in any current or future media, including reprinting/republishing this material for advertising or promotional purposes, creating new collective works, for resale or redistribution to servers or lists, or reuse of any copyrighted component of this work in other works.} \\ \\
	
	{\Large Published in \emph{IEEE Access}, 29 January 2026.} \\ \\

	Cite as:
	
	\vspace{0.1cm}
	\noindent\fbox{%
		\parbox{\textwidth}{%
			L.~Ullrich, M.~Buchholz, K.~Dietmayer, and K.~Graichen, "Toward Fully Autonomous Driving: AI, Challenges, Opportunities, and Needs," in \emph{IEEE Access}, 29 January 2026, pp. 1--26, doi: 10.1109/ACCESS.2026.3659192.
		}%
	}
	\vspace{2cm}
	
\end{@twocolumnfalse}
]

\noindent\begin{minipage}{\textwidth}
	
\hologo{BibTeX}:
\footnotesize
\begin{lstlisting}[frame=single]
@article{ullrich2024adstack,
	title={Toward Fully Autonomous Driving: AI, Challenges, Opportunities, and Needs},
	author={Ullrich, Lars and Buchholz, Michael and Dietmayer, Klaus and Graichen, Knut},
	journal={IEEE Access},
	year={2026},
	pages={1--26},
	doi={10.1109/ACCESS.2026.3659192},
	publisher={IEEE}
}
\end{lstlisting}
\end{minipage}
\setcounter{page}{0}

	
\bstctlcite{IEEEexample:BSTcontrol}
\history{
	Received 1 December 2025, accepted 26 January 2026.}
\doi{10.1109/ACCESS.2026.3659192}


\title{Toward Fully Autonomous Driving: \\AI, Challenges, Opportunities, and Needs}
\author{\uppercase{Lars Ullrich}$^{\orcidicon{0009-0001-8166-3118}}$\authorrefmark{1},
\uppercase{Michael Buchholz}$^{\orcidicon{0000-0001-5973-0794}}$\authorrefmark{2}, \uppercase{Klaus Dietmayer}$^{\orcidicon{0000-0002-1651-014X}}$\authorrefmark{2},
\IEEEmembership{Senior Member,~IEEE}, 
and \uppercase{Knut Graichen}$^{\orcidicon{0000-0003-2865-8093}}$\authorrefmark{1},
\IEEEmembership{Senior Member,~IEEE}}

\address[1]{Chair of Automatic Control, Friedrich-Alexander-Universität Erlangen-Nürnberg (FAU), 91058 Erlangen, Germany}
\address[2]{Institute of Measurement, Control and Microtechnology, Ulm University, 89081 Ulm, Germany}
\tfootnote{This research is accomplished within the project ”AUTOtech.agil” (FKZ 01IS22088Y, FKZ 01IS22088W). We acknowledge the financial support for the project by the Federal Ministry of Education and Research of Germany (BMBF).}

\markboth
{Ullrich \headeretal: Toward Fully Autonomous Driving: AI, Challenges, Opportunities, and Needs}
{Ullrich \headeretal: Toward Fully Autonomous Driving: AI, Challenges, Opportunities, and Needs}

\corresp{Corresponding author: Lars Ullrich (e-mail: lars.ullrich@fau.de).}

\begin{abstract}
Efficient scalability of automated driving (AD) is key to reducing costs, enhancing safety, conserving resources, and maximizing impact. However, research focuses on specific vehicles and context, while broad deployment requires scalability across various configurations and environments. Differences in vehicle types, sensors, actuators, but also traffic regulations, legal requirements, cultural dynamics, or even ethical paradigms demand high flexibility of data-driven developed capabilities. In this paper, we address the challenge of scalable adaptation of generic capabilities to desired systems and environments. Our concept follows a two-stage fine-tuning process. In the first stage, fine-tuning to the specific environment takes place through a country-specific reward model that serves as an interface between technological adaptations and socio-political requirements. In the second stage, vehicle-specific transfer learning facilitates system adaptation and governs the validation of design decisions. In sum, our concept offers a data-driven process that integrates both technological and socio-political aspects, enabling effective scalability across technical, legal, cultural, and ethical differences.
\end{abstract}

\begin{abstract}\label{00_Abstract}
	Automated driving (AD) is promising, but the transition to fully autonomous driving is, among other things, subject to the real, ever-changing open world and the resulting challenges. However, research in the field of AD demonstrates the ability of artificial intelligence (AI) to outperform classical approaches, handle higher complexities, and reach a new level of autonomy. At the same time, the use of AI raises further questions of safety and transferability. To identify the challenges and opportunities arising from AI concerning autonomous driving functionalities, we have analyzed the current state of AD, outlined limitations, and identified foreseeable technological possibilities. Thereby, various further challenges are examined in the context of prospective developments. In this way, this article reconsiders fully autonomous driving with respect to advancements in the field of AI and carves out the respective needs and resulting research questions.
\end{abstract}


\titlepgskip=-21pt

\maketitle

\section{INTRODUCTION}\label{1}

Automated driving (AD) represents a well-known safety-critical application that operates in the real world environment striving for safer and more efficient mobility \cite{crayton2017autonomous}. While several simple driver assistance systems do not require the use of artificial intelligence (AI) \cite{liebemann2004safety, winner2009adaptive}, it is becoming increasingly important in highly automated vehicles \cite{gupta2021deep, guo2022review, claussmann2017study, leurent2020safe, filos2020can}. This is due to the fact that increasing responsibility for operational processes in complex and changing environments leads to more demanding decision-making \cite{canas2003cognitive}. In particular, the inherent complexity of the task itself, the uncertainty in the predictability of other road users due to the emergent behavior, and the complexity of interactions of various local and decentralized technical systems with non-technical participants pose a tremendous challenge \cite{burton2023addressing}.

Currently, the complexity of AD is primarily handled by a modular, service-oriented software architecture \cite{hellmund2016robot, kugele2017service}, where AI is increasingly being used in the individual sub-modules, such as perception \cite{horn2020deepclr} or planning \cite{claussmann2019review}. This generally enables an increase in the performance of the individual modules, but raises questions about safety \cite{neto2022safety} and explainability \cite{kuznietsov2024explainable}. This field of tension between improved performance and growing concerns regarding safety and explainability is further amplified in fully monolithic, AI-only architectures, where the absence of modular validation and interpretability layers exacerbates these challenges \cite{chen2023end}.

At the same time, AI methodology is increasingly evolving. In particular, new learning techniques are promising, such as zero-shot \cite{pmlr-v37-romera-paredes15}, \cite{10.1145/3293318}, one-shot \cite{fei2006one, vinyals2016matching}, few-shot \cite{wang2020generalizing}, \cite{kadam2020review}, and meta-learning \cite{hochreiter2001learning, finn2017model, vanschoren2019meta}. These techniques are especially relevant with respect to generalization, which is crucial for the application of AI in the real world \cite{dubey2021adaptive}, \cite{maharaj2022generalizing}. In addition, foundation models (FMs) show tremendous performance in areas such as  language \cite{openai2023gpt4, touvron2023llama} and vision \cite{chowdhery2022palm, kirillov2023segment, Liu_2022_CVPR}. Moreover, FM offer new opportunities in scenarios engineering (SE) \cite{li2022novel, wang2024does}, but also pose new challenges \cite{raffel2020exploring, floridi2020gpt, kasneci2023chatgpt}. 

Accordingly, it is necessary to reconsider the current state of AD with regard to the aim of fully autonomous driving. This is in particular relevant in response to existing changes in technological developments and possibilities. This paper is dedicated to this task and therefore contributes in the following respects: 
\begin{enumerate}
	\item A critical examination of the current state of AD is presented, focusing on current limitations of AD towards fully autonomous driving.
	\item An analysis of the necessary steps to achieve higher autonomy and advance towards fully autonomous driving is conducted, considering emerging technologies, particularly AI, and identifying potential developments.
	\item An extended analysis of the challenges associated with the widespread deployment of autonomous driving is conducted, going beyond purely functional aspects. This extended analysis identifies potential countermeasures necessary for scalable deployment across various vehicle configurations and environments,  while opening up new research questions.
\end{enumerate}

The aim of the paper is to outline the current status and limitations of AD systems, highlighting the need for adaptation. While emphasizing functionality, related aspects such as scalability are also considered. The focus is on analyzing the challenges and opportunities of AI for fully autonomous driving, alongside exploring emerging needs and new perspectives. To achieve this, the methodology combines a qualitative, knowledge-driven literature review with forward-looking perspectives, incorporating novel insights and conceptual frameworks to address prospective opportunities, challenges, and emerging needs.

The paper is structured as follows: AD and the current limitations are described in Section \ref{AutomatedDriving}. Building on this, possibilities towards fully autonomous driving are outlined in Section \ref{AutonomousDriving}. In this regard, prospective challenges and opportunities are identified and outlined in Section \ref{FutureAD}. Finally, Section \ref{ADDiscussion} presents the discussion, followed by the conclusion in Section \ref{ADConclusion}

\section{Automated Driving and Current Limitations}\label{AutomatedDriving}
This section is dedicated to AD, the state of the art (SOTA) tech stack and the increasing use of AI. In addition, emerging research areas and corresponding challenges concerning fully autonomous driving are discussed. 

\subsection{Modular Service-Oriented Software Architecture \& AI}\label{Modular_ASOA}

The current SOTA AD stack is
predominantly characterized by a modular, service-oriented
software architecture and the increasing use of AI subsystems.
This is illustrated in more detail in the following.

\subsubsection{SOTA Architecture}
\begin{figure}[h]
	\centering	
	\includegraphics[width=\linewidth]{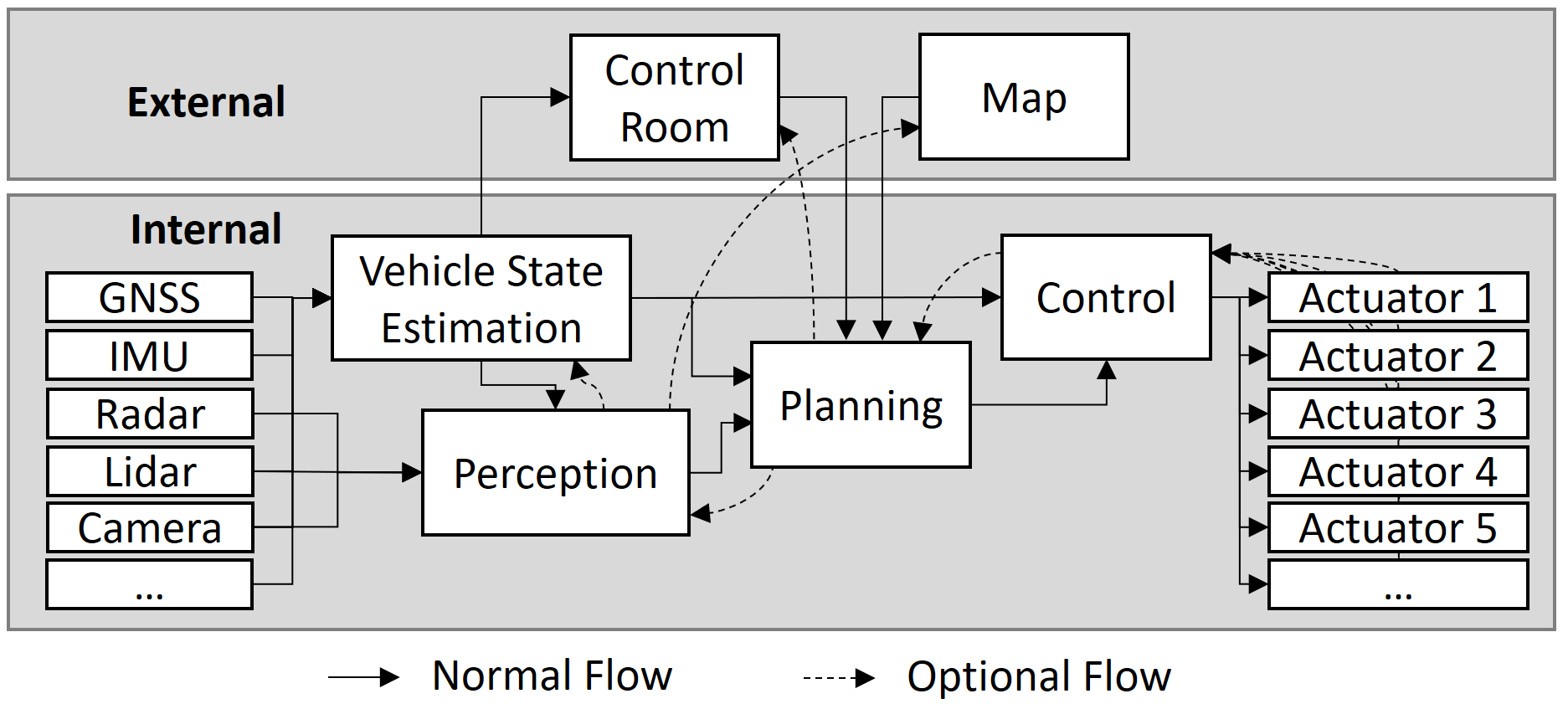}
	\caption{High-level illustration of an exemplary modular service-oriented software architecture for AD on the sensor/actor level.}
	\label{fig:architecture}
\end{figure}

Current approaches to manage complexity are often based on a modular and service-oriented information processing chain \cite{hellmund2016robot, kugele2017service, kampmann2019dynamic, becker2021safety, zhu2021requirements}. Individual modules encapsulate functionalities and responsibilities such as perception or planning. These higher-level logically separated functions are ultimately implemented by a large number of various sub-modules or services. Figure \ref{fig:architecture} shows an example of the modular service-oriented AD architecture for the core functions between sensors and actuators. Perception, for instance, collects lidar, radar, and camera data as well as pre-processed vehicle state estimation data in order to subsequently generate an environment model, which may consist of free space and occupancy maps. 

\subsubsection{SOTA AI Usage}
In individual sub-modules, research on AI has been conducted in a targeted manner. In particular, publicly available datasets \cite{geiger2013vision, huang2019apolloscape, caesar2020nuscenes, sun2020scalability, chang2019argoverse, zhan2019interaction, malinin2021shifts, houston2021one, caesar2021nuplan} and associated challenges make an important contribution to researching and comparing different approaches, e.g., in the field of perception \cite{horn2020deepclr}, in object detection \cite{wirges2019capturing, feng2020deep, gupta2021deep}, and in object tracking \cite{Luo_2021_ICCV, chiu2021probabilistic, guo2022review}. But also in the area of planning \cite{claussmann2019review}, more precisely in the area of prediction of other road users \cite{ding20201st, varadarajan2022multipath++, liu2022strajnet, casas2020implicit}, AI methods are increasingly applied. Especially the superior performance in terms of accuracy compared to traditional non-AI methods justifies their exploration \cite{grigorescu2020survey, ma2020artificial, karle2022scenario}. Even though improved performance provides greater safety, the lack of validation of methods increases the risk of catastrophic consequences when humans are no longer in the loop. While the recognition of traffic signs in road traffic already works via AI systems such as \cite{stallkamp2012man, tabernik2019deep} and presents the corresponding information to the driver via head-up displays \cite{abdi2017deep}, the functionality does not yet decisively take control of the system but merely informs the driver \cite{murugan2022autonomous}, who processes the information on his own responsibility. 

\subsubsection{Hurdle of AI Usage}
If AI components are used in a highly automated or even fully autonomous vehicles, it must be ensured that either no errors occur at all time or that a monitoring component detects errors and prevents catastrophic consequences by intervening accordingly \cite{cunneen2019autonomous, taeihagh2021governance}. Consequently, ensuring safety is a key aspect in order to ultimately use AI methods without risk, thus ensuring the overarching goal of fully autonomous driving--safer traffic.  

\subsubsection{Related Safety Assurance}
Securing and guaranteeing the correct functionality takes up the current trend of modular service-oriented architectures \cite{dajsuren2019automotive, gomez2022build}. Taking into account hard-coded interfaces between services, it is possible to secure individual services independently \cite{philion2020learning, klamann2023introducing}. As a result, it is possible to update and upgrade services without having to secure the entire system again \cite{stolte2020towards, guissouma2022lifecycle}. This requires that the respective service performs according to its specifications. 

In terms of AI safety assurance, existing challenges \cite{amodei2016concrete, tabani2019assessing} need to be addressed. As shown in \cite{neto2022safety}, the importance of this research domain has increased significantly in recent years but still poses various open questions. Due to the fact that other articles deal with the analysis and discussion of AI assurance \cite{batarseh2021survey}, trustworthiness \cite{diaz2023connecting}, and governance \cite{dafoe2018ai} in general, but also the AI safety assurance \cite{ullrich2024aisafety_assurance} and the explainability \cite{atakishiyev2024explainable, kuznietsov2024explainable, atakishiyev2024safety} with respect to AD, we refer to the corresponding literature for these aspects, while focusing on the functionality. Accordingly, the relevant function-specific research areas are reviewed and the prospective challenges towards fully autonomous driving are discussed within this paper. 

\subsection{Situation Awareness for Autonomy Advancements}\label{2b}

While the AD stack and the increasing use of AI within the AD stack were discussed previously, this subsection deals with situation awareness (SA), which is of particular importance for achieving improvements in autonomy. Hereby, while SA is broadly relevant, e.g., for HMI \cite{atakishiyev2024incorporating} and explainable AI \cite{sanneman2022situation}, the following concentrates on the role of SA within the AD stack and driving functionalities, in line with the focus of this article.

\subsubsection{General Introduction} SA is taken up according to the most widespread definition of \cite{endsley1988design, endsley1995toward}, stating that SA comprises three levels: the perception of the environment (Level 1 SA), the comprehension of meaning (Level 2 SA), and the projection of the future (Level 3 SA) and forms the basis of subsequent decision-making. Consequently, in the chain of sensing, perception, planning, control, and actuation, SA represents the cognitive link between perception and planning, which is important for the progress towards fully autonomous driving by increasing the functional responsibility for operational processes. Accordingly, SA comprises a functional sub-area within the AD stack that is less well settled compared to sensing or trajectory tracking. Therefore, SA is a key area of current research \cite{sirkin2017toward, gerwien2021towards, ignatious2023analyzing} and exhibits considerable research questions. In line with this, research in terms of functionality is focusing on increasingly complex situations and the associated demand for greater cognition, which is in turn stimulating the increasing deployment of high-performing AI systems in actual applications, e.g., \cite{romera2017erfnet, wischnewski2022indy, wang2023yolov7, munir2022situational, ravanbakhsh2018learning}. 

Due to the fact that Level 1 SA is represented by perception, and Level 2 SA, comprehension, directly builds on Level 1 SA, SA has always been strongly linked to perception from a functional point of view. This is why SA is sometimes regarded as the further development of traditional perception to active perception, also referred to as situation-aware environment perception \cite{henning2022situation}. However, the comprehension of meaning and significance of Level 2 SA goes beyond a situation-aware environment perception and Level 3 SA, projection, even reaches profoundly into planning  \cite{endsley2020situation}. Indeed, the projection, also called prediction, e.g., of other road users \cite{huang2022survey} and consecutive projection of a collision likelihood \cite{tolksdorf2024fast} within predictive planning \cite{althoff2009model, schwarting2017safe}, is in turn a central component of trajectory planning. As a result, it is clear that the SA is not an individual, logically separated functional module of the AD stack, but rather spans functionalities and responsibilities across the AD stack. The current status is discussed in more detail below based on the individual SA Levels. 

\subsubsection{Level 1 SA} This stage covers classic perception topics from object detection  \cite{wirges2019capturing, feng2020deep, gupta2021deep}, to object tracking \cite{Luo_2021_ICCV, chiu2021probabilistic, guo2022review} as well as the acquisition of daytime, weather conditions and data provided by the environment \cite{endsley2020situation}. While appropriate hardware enables multi-modal, multi-redundant sensor systems to provide a comprehensive environmental perception \cite{liu2020computing}, current research is concerned with real-time capability and resource efficiency. Therefore, initial approaches were based on sequential perception and subsequent filtering using heuristics, such as the proximity of objects \cite{bajcsy2018revisiting, rasouli2020attention, Pal_2020_CVPR}, or a-priori contexts \cite{xu2015context, nager2019lies}. More recent approaches such as \cite{henning2022situation}, on the other hand, rely on a high-level situation analysis, which in turn directly influences perception processing and thus actively directs perception. In addition to situation-aware environment perception approaches \cite{sirkin2017toward, gerwien2021towards, henning2022situation, ignatious2023analyzing, cheng2022masked, li2022bevformer, liu2023bevfusion, zhang2023occformer, huang2023tri, wei2024bev}, which demonstrate an increasing fusion of Level 1 SA and Level 2 SA, new concepts are also being explored through hardware. For example, event-based neuromorphic vision \cite{chen2020event}, which already incorporates a paradigm shift in the hardware from continuous to event-triggered perception processing and thus reduces the resource demand without requiring a situation analysis. 

\begin{figure}
	\centering	
	\begin{subfigure}{0.46\textwidth}
		\includegraphics[width=\linewidth]{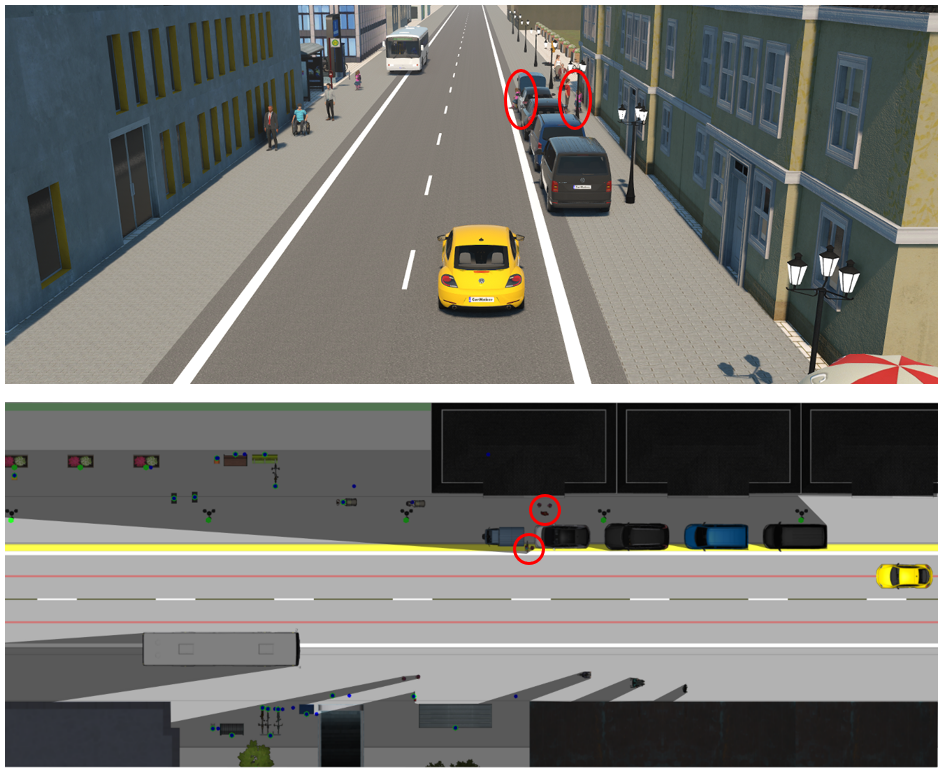}
		\caption{Urban driving situation with schoolchildren (red circled) intending to cross the road. The plain perception-based free space is shown from the bird's eye view (BEV) perspective.}
		\label{fig:freenotfreeschool}
	\end{subfigure}
	\begin{subfigure}{0.46\textwidth}
		\includegraphics[width=\linewidth]{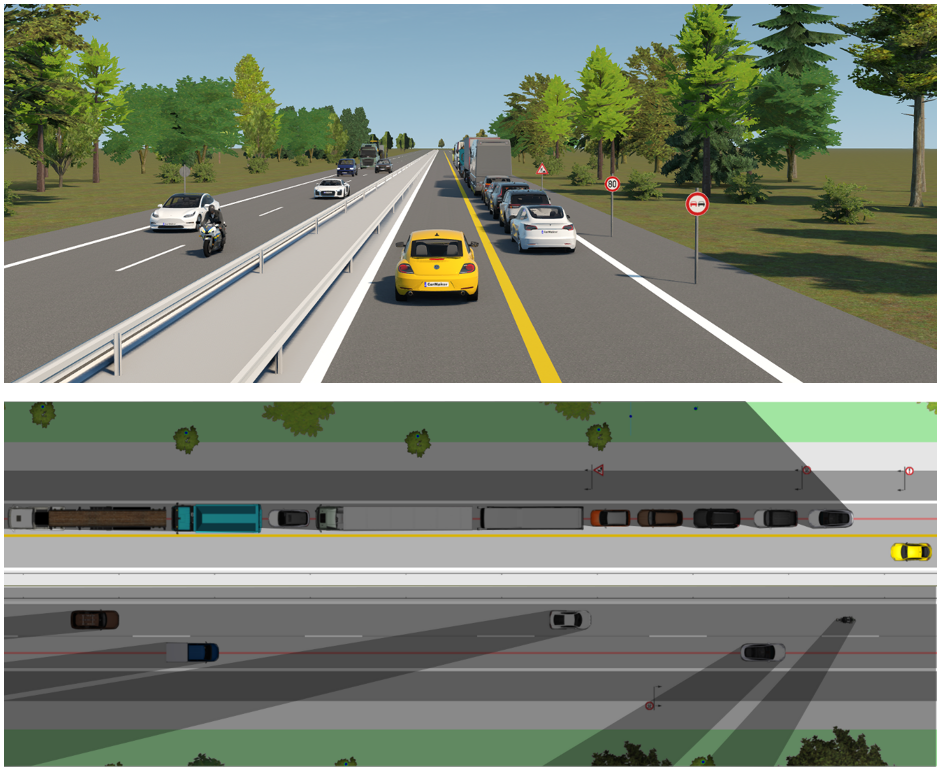}
		\caption{Highway driving situation with a congestion on the right lane. The plain perception-based free space is shown from the BEV perspective.}
		\label{fig:freenotfreehighway}
	\end{subfigure}
	\caption{Comparison of different driving situations with comparably perceived free space. In case (a), the context of the situation allows the free space to be used. In situation (b), however, caution is advised. The differences result from the context. This illustrates that not all free space is equal.}
	\label{fig:freenot}
\end{figure}

\subsubsection{Level 2 SA} This stage considers the comprehension of meaning and significance, and, thus represents the transition from perceiving to understanding the environment \cite{endsley2020situation}. In this regard, comprehension builds on perception by interpreting relationships between entities in the environment, which according to \cite{ulbrich2015defining}, is also referred to as scene modeling. Following the definitions of \cite{ulbrich2015defining}, a scene is transferred to a situation through pattern recognition, interpretation and goal and value consideration, leading to the entirety of circumstances to be reflected in subsequent behavior and motion planning. The major challenge for autonomous driving in this context is to achieve efficient, real-time capable, effective and nuanced comprehension that is able to process the numerous edge cases in a reasonable way. In this context, capabilities according to \cite{koopman2016challenges, koopman2017autonomous} are limited.  To illustrate the complexity, consider the scenes depicted in Figure \ref{fig:freenot}. In both cases, the perception-based free space is highlighted. However, it is the responsibility of Level 2 SA to discern the nuances between the two situations. In Figure \ref{fig:freenotfreehighway}, the free space is drivable, whereas in Figure \ref{fig:freenotfreeschool}, the free space is not unconditionally drivable. Level 2 SA must understand that in Figure \ref{fig:freenotfreeschool}, there is a school on the right-hand side and a school bus waiting on the left, at a time when school has just let out. This suggests that children may be crossing between parked vehicles on the right side. Consequently, the maximum allowable speed might not be appropriate. In contrast, Figure \ref{fig:freenotfreehighway} presents a situation where, according to the applicable traffic rules, the free space can reasonably be considered unconditionally drivable. Although more recent approaches such as \cite{sirkin2017toward, gerwien2021towards, henning2022situation, ignatious2023analyzing} are beginning to address related concerns, the human driver is still superior at integrating and nuancing all relevant information. For instance, in the scenario in Figure \ref{fig:freenotfreeschool}, whether it is a weekday or a holiday can have a significant impact. In this respect, AI is a promising way to improve capabilities. However, while the above example highlights a specific situation, countless other difficult situations reflect the complexity of achieving a reliable level 2 SA. 

\subsubsection{Level 3 SA} This stage deals with the projection into the future. Due to the emergent behavior of other road users, it is not possible to make statements with certainty \cite{endsley2020situation}. The goals of other road users and their respective behavioral decisions are also not known with certainty. For this reason, a variety of motion prediction approaches have been developed, as outlined in \cite{gulzar2021survey}. While learning-based multi-modal joint predictions, that are interaction-aware, are currently widely used \cite{trentin2023multi, shi2024mtrpp, gan2023mgtr}, other approaches are still emerging, e.g., occupancy flow \cite{mahjourian2022occupancy, agro2023implicit, tong2023scene, liulet}. Nevertheless, current research is particularly concerned with the appropriate integration of prediction into planning \cite{hagedorn2023rethinking}. Traditionally, prediction and planning are sequential tasks. Depending on the conditioning of the predictions, the ego-vehicle behaves more or less aggressively. While predictions conditioned to the ego-plan lead to more aggressive behaviors \cite{tang2019multiple, schmerling2018multimodal}, plans based on predictions that are unconditioned to the ego-plan are inherently more reactive and less aggressive \cite{zeng2019end, sadat2020perceive}. Therefore, increasing efforts are being made \cite{sun2022m2i, rhinehart2021contingencies} to overcome the sequential prediction-planning process in favor of an enhanced bi-directional interaction between prediction and planning to achieve neither aggressive nor overly passive behaviors.

\subsubsection{Conclusion}

It can be seen that the perception of the environment (Level 1 SA) can be improved by means of a high-level situation analysis (Level 2 SA, Level 3 SA). At the same time, it is evident that the comprehension of meaning (Level 2 SA) is largely based on the perception of the environment (Level 1 SA) and can benefit from a projection of the future (Level 3 SA). The projection of the future (Level 3 SA) is in turn highly dependent on a comprehensive and nuanced understanding of the situation (Level 2 SA) and thus indirectly on the perception (Level 1 SA). Furthermore, it can be recognized that the decision-making in terms of trajectory planning is also increasingly bi-directionally affected by the prediction of other road users (Level 3 SA). Overall, it is therefore apparent that the individual levels of SA and, with regard to the AD stack, the individual modules and functionalities have recognizable dependencies in the pursuit of the desired autonomy. 

However, a demanding consequence of these observations and developments is the increasing circularity, which is also the main concern of Endsley's SA model \cite{flach1995situation}. This in turn means that with increasing bi-directionality and interdependency, modular safeguarding  \cite{philion2020learning, klamann2023introducing} reaches its limits, as interfaces and passed values are permissible, but circular reasoning may lead to undesirable behavior. Beyond this, decision-making requirements also arise with increasing SA. For instance, in order to be able to react appropriately to the nuanced SA, it is necessary to move away from decision trees and rule-based "if this, then that" structures within decision-making in order to deal with the diversity and complexity of the real world \cite{chen2023end}.

\begin{figure}[]
	\centering	
	\vspace*{1mm}
	\includegraphics[scale=0.38]{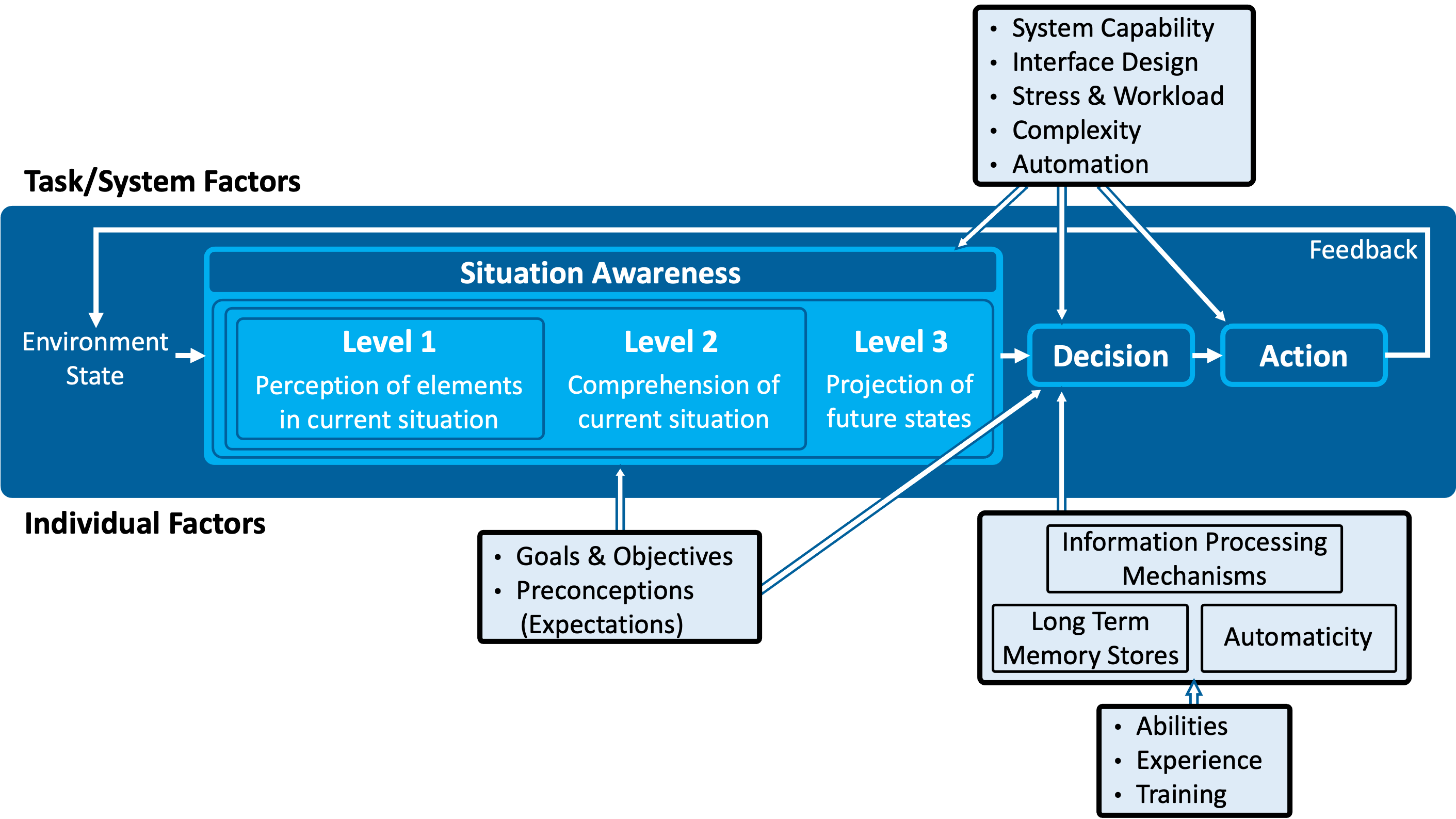}
	\caption{Endsley's model of situation awareness in dynamic decision making according to \cite{endsley1995toward}.}
	\label{fig:SA_Endsley}
\end{figure}

To sum up, the SA-driven limitation analysis provides indications for further developments alongside foreseeable challenges. Thereby, in particular, the theory of SA in dynamic systems \cite{endsley1995toward}, illustrated in Figure \ref{fig:SA_Endsley}, enables an even broader perspective of analysis. In particular, with regard to individual factors such as information processing mechanisms and individual abilities, or external/environmental factors such as task complexity and system-environment interfaces. While within this subsection the SA and the individual factors, e.g., the need for flexible information processing mechanisms \cite{kim2012safer, schlatow2015extensible, kampmann2019dynamic} have been considered with respect to SOTA limitations, the following section will broaden the perspective of limitation analysis towards external factors. 

\subsection{Increased Interconnectivity Mitigates Complexity}

One way to address the complexity of SA in dynamic decision-making, Figure \ref{fig:SA_Endsley}, and autonomous driving in general, is offered by increasing interconnectivity and collaboration \cite{wang2018networking, bagheri20215g}. Joint fused environment models \cite{lampe2020reducing, godoy2021grid, volk2021towards} enable information to be extracted from areas that are not individually visible. In this context, mobile-edge computing \cite{zhang2017mobile} offers the possibility to efficiently create infrastructure environment models, that can complement the vehicle's perception to extend the vehicle's field of view (FOV) and improve the resulting behavior, as demonstrated in \cite{buchholz2021handling}. Furthermore, while traffic surveillance \cite{datondji2016survey} and traffic analysis \cite{fleck2019towards} could be a preliminary stage in this process, but also for the uncertainty reduction in future projection, shared goals and trajectories offer significant uncertainty and complexity reduction opportunities, thus enabling more efficient operations \cite{lampe2022cloud}. Decentralized cooperative behavior and trajectory planning reinforce this further \cite{viana2019cooperative, burger2020interaction, yurtsever2020survey}. At the same time, this requires decentralized computing capacity \cite{lampe2022cloud} and raises new challenges. Besides technical difficulties like latency, cyber security is a key aspect in this regard \cite{alnasser2019cyber, kim2021cybersecurity}. Communication with road side units as well as the closure of safety contracts with infrastructure reduce the danger of misinterpretations and cut the complexity \cite{arechiga2019specifying, grobelna2021dynamic, zacchi2021towards}. 

Although these methods offer undeniable advantages, they require additional resources and the renewal of the infrastructure. Moreover, the difficulty of interpreting and predicting other road users, e.g., pedestrians in collaboratively shared spaces, remains, as a disruptive step towards purely autonomous vehicles is neither foreseeable nor targeted. In addition, independent interpretation and decision-making is an essential safety factor and a crucial fallback component in the event of malfunctions in infrastructure or decentralized services. Accordingly, while the research area of increased interconnectivity contributes to mitigate complexity, it does not provide a comprehensive solution to the task-inherent complexity. Overall, interconnectivity provides extensive information to improve SA and decision-making, but also increases the challenge of focusing on and extracting the most important information. This is another area where AI could be promising to process the multitude of external information and target it according to context-specific relevance.

\subsection{Semantic Gap Challenges Operation}

\begin{figure}[]
	\centering	
	\includegraphics[scale=0.55]{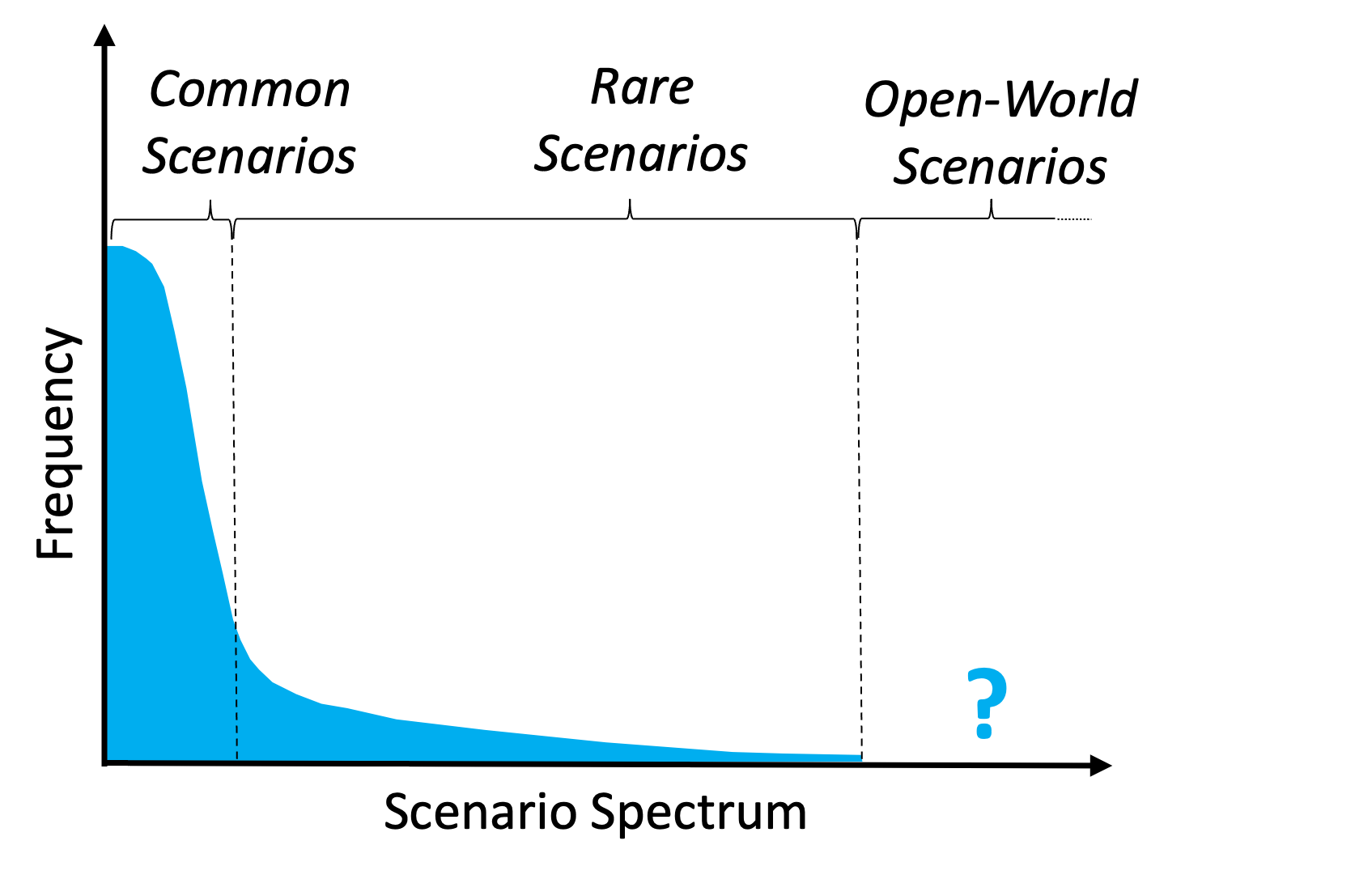}
	\caption{The spectrum of scenarios in autonomous driving in the real world is described as an open long-tail distribution. In addition to scenarios that occur very frequently, there are a large number of rare scenarios and the challenge of novel scenarios that arise due to the characteristics of the open world.}
	\label{fig:cornercase}
\end{figure}

A more general challenge within AD is the semantic gap, namely the discrepancy between implied system functionality and realized system functionality \cite{burton2020mind}, that becomes particularly visible during deployment. Identified gaps are usually eliminated by extending the functional scope. New functionalities are implemented through additional services or operating modes \cite{ecksteinautotech}. However, numerous corner cases and the continuous changes in the open world lead to an open long-tail distribution \cite{liu2019large, chen2023end}, as illustrated in Figure \ref{fig:cornercase}. This poses particular challenges for the resilience and robustness of the systems. Currently, this issue is emphasized by the increasing deployment of automated vehicles in the USA. Everyday situations, which are handled by humans, can cause particular difficulties. For instance, automated vehicles get stuck in concrete under construction \cite{templeton2023cruise}, hit a pedestrian in a complex situation \cite{NHTSARecall23E}, or fail to clear the way for emergency vehicles \cite{californiaProbe2023}. 

Since the obstruction of emergency vehicles is a hazard to the general public, addressing this problem is of significant relevance. Functionalities such as machine listening, which recognize siren noises including the direction of origin and distinguish them from fakes, are being developed \cite{keutgens2019soundai, kwade2021hearingsense}. While such functionalities, or also dedicated intercommunication, address the emergency clearance, it does not provide a solution for other challenging situations, e.g., the concrete under construction failure \cite{templeton2023cruise}. Through the dedicated handling of critical use cases, automated vehicles are becoming increasingly complex systems of systems, with expertise being induced into the architecture via the dedicated solutions and respective integrations. This counteracts the flexibility required to handle the open long-tail distribution of corner cases, as the semantic gap is not generally addressed. Accordingly, the current methodology contradicts the human approach to cope with complex situations through dynamic decision-making, cognitive flexibility and adaptability to the environment and its changes \cite{canas2003cognitive, brehmer1992dynamic}, especially at the core of the cognitive processes from perception to planning. 

Due to the fact that the necessary level of cognition has not yet been achieved, there exist approaches for resolving challenging situations by means of infrastructure or even external support, e.g., such as fallback teleoperation via control rooms \cite{feiler2020concept}, to accomplish AD missions. Nevertheless, fully autonomous driving aims to address the open-world scenarios through a universal approach. In this regard, AI and the generalization of AI could provide a remedy.

\subsection{(Modular) End-to-End Learning}

The existing limitations of SOTA modular AD stacks \cite{urmson2008autonomous, levinson2011towards, wei2013towards, ziegler2014making, hellmund2016robot, kugele2017service, paden2016survey, yurtsever2020survey}, e.g., in SA and cognition in general, and especially the emerging capabilities of AI motivated researching the use of AI across the overall AD Stack at once, referred to as end-to-end (E2E) AD architectures \cite{pomerleau1988alvinn, bojarski2016end, sallab2017deep, codevilla2018end, codevilla2019exploring, hawke2020urban, chen2020learning, chen2021learning, zhang2022rethinking, hu2022model}. In this regard, the recently popular FMs, such as large language models (LLM) \cite{ray2023chatgpt, kitaev2018multilingual}, vision-language models (VLM) \cite{zhu2023minigpt, liu2024visual}, and vision-language-action models (VLA) \cite{zitkovich2023rt, kim2024openvla} are also receiving more attention within AD, from vision \cite{zhou2024vision} to motion planning \cite{seff2023motionlm} and E2E AD \cite{cui2023drivellm, sima2023drivelm, xu2024drivegpt4, you2024v2x, tian2024drivevlm, zhou2025opendrivevla} in general. Overall, E2E AD architectures counteract compounded errors, objective mismatches, and information losses \cite{luo2018fast, liang2020pnpnet, sadat2020perceive, weng2022mtp, weng2022whose, eysenbach2022mismatched} of modular information processing chains and enable improved performance. Nevertheless, E2E AD stacks have significant drawbacks, e.g., the loss of interpretability, which raises concerns regarding debugging, verification, and safety guarantees. Furthermore, initial E2E AD stacks are more sensitive to distribution shifts \cite{casas2021mp3, codevilla2019exploring, ross2011reduction}. For this reasons and the respective open research questions of AI explainability \cite{atakishiyev2024explainable, kuznietsov2024explainable}, trustworthiness \cite{diaz2023connecting}, and safety assurance \cite{ullrich2024aisafety_assurance}, modular AD stacks are still predominantly deployed \cite{uber2020safetyreport, motional2021selfassessment, waymo2021safetyreport}. 

\begin{figure*}
    \centering
    \begin{tikzpicture}
        \node[anchor=south west, inner sep=0] (image) at (-0.36,-0.36) 
            {\includegraphics[width=\textwidth]{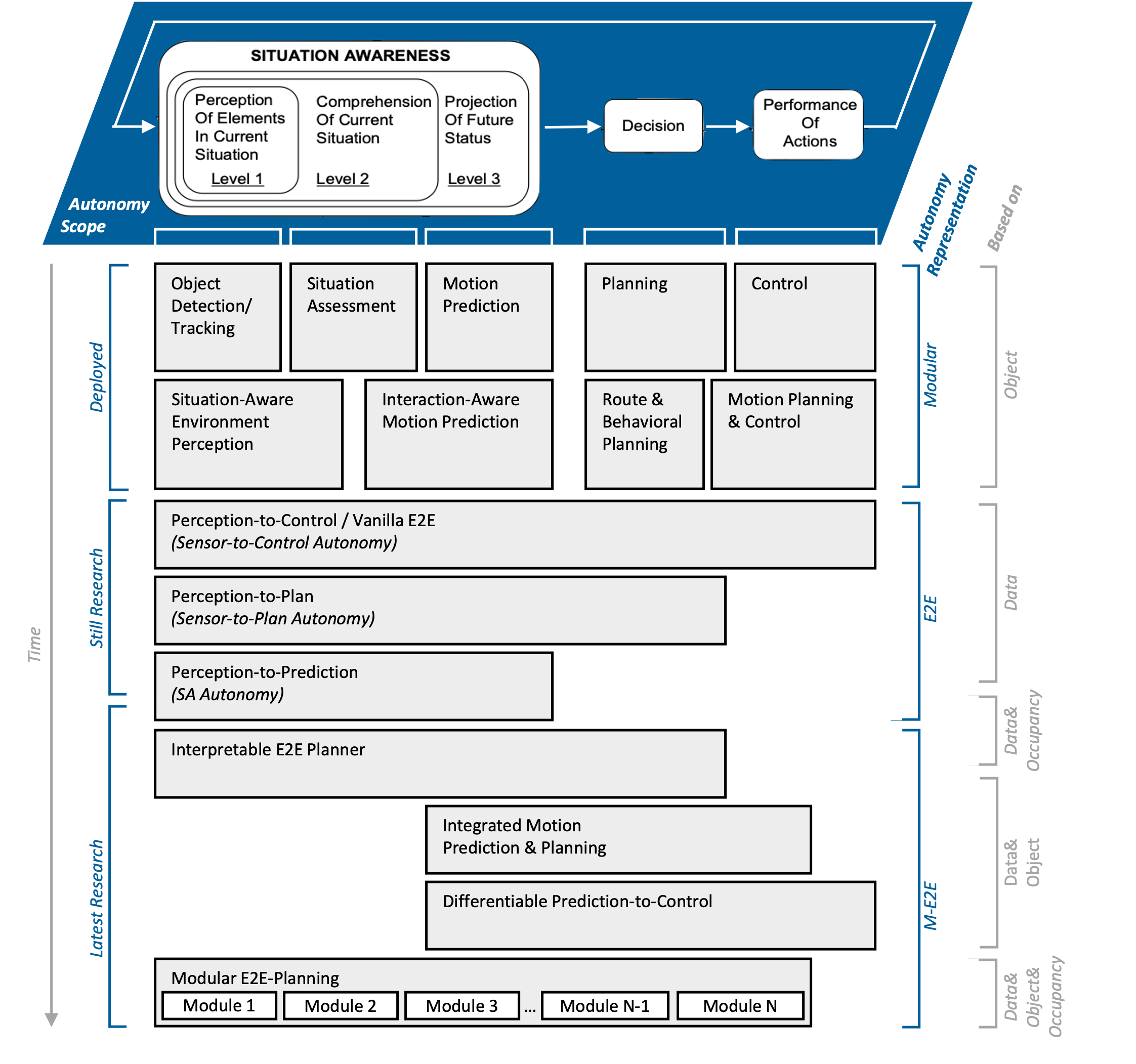}};

        \node[anchor=north west] at (1.99,11.1){\cite{carion2020end, zeng2022motr}};
        \node[anchor=north west] at (4.16,11.45){\cite{ulbrich2015situation, scheel2022recurrent}};
        \node[anchor=north west] at (6.25,11.46){\cite{ casas2018intentnet, phan2020covernet}};
        \node[anchor=north west] at (8.87,11.8){\cite{ lim2018hierarchical,sadat2019jointly},};
        \node[anchor=north west] at (8.87,11.4){\cite{chen2023tree, vitelli2022safetynet, mao2023gpt}};

        \node[anchor=north west] at
        (11.21,11.8){\cite{yoon2009model, ji2018adaptive}};

        \node[anchor=north west] at (3.77,10.01){\cite{feng2020deep},};
        \node[anchor=north west] at (3.77,9.65){\cite{henning2022situation},};
        \node[anchor=north west] at (2.07,9.28){\cite{ casas2018intentnet, cheng2022masked, li2022bevformer, liu2023bevfusion, zhang2023occformer, huang2023tri, wei2024bev}};
        \node[anchor=north west] at (5.28,9.63){\cite{
        varadarajan2022multipath++, casas2020implicit, liu2022strajnet, trentin2023multi},};
        \node[anchor=north west] at (5.28,9.25){\cite{shi2022motion, lee2017desire, ngiamscene, jia2023hdgt, seff2023motionlm}};
        \node[anchor=north west] at (8.69,9.28){\cite{wei2014behavioral, orzechowski2020decision}};
        \node[anchor=north west] at (10.85,9.63){\cite{li2022learning, menner2020inverse, bae2022lane, sha2023languagempc}};

        \node[anchor=north west] at (6.31,8.46){\cite{ bojarski2016end, sallab2017deep, codevilla2018end, codevilla2019exploring, hawke2020urban, toromanoff2020end, Rhinehart2020Deep,
        ohn2020learning, chen2020learning, chen2021learning,
        zhang2021end,
        wu2022trajectory,
        scheel2022urban,
        zhang2022rethinking,
        jia2023think,
        lihydra}};
        
        \node[anchor=north west] at (4.46,7.25){\cite{ pomerleau1988alvinn, mueller18a, prakash2021multi, chitta2022transfuser, renz2022plant, jia2023driveadapter, jia2023driveadapter, wang2023drivemlm}};

        \node[anchor=north west] at (5.17,6.09){\cite{liang2020pnpnet,hu2021fiery, khurana2022differentiable},};
        \node[anchor=north west] at (4.19,5.72){\cite{ gu2023vip3d, mahjourian2022occupancy, agro2023implicit, tong2023scene, liulet, li2025viewformer}};

        \node[anchor=north west] at (5.19,4.87){\cite{zeng2019end,sadat2020perceive, casas2021mp3, hu2022model, biswas2024quad, wang2025alpamayo},};
        \node[anchor=north west] at (5.19,4.47){\cite{cui2021lookout, zeng2020dsdnet, chitta2021neat, chen2022learning, philion2020lift, hu2022st, shao2023safety, cui2023drivellm, sima2023drivelm, xu2024drivegpt4, you2024v2x, tian2024drivevlm, chen2024driving}};

        \node[anchor=north west] at (9.04,3.67){\cite{song2020pip, hu2021safe, pini2023safe, chen2023interactive},};
        \node[anchor=north west] at (9.04,3.30){\cite{ huang2023gameformer, huang2024dtpp}};

        \node[anchor=north west] at (10.72,2.51){\cite{huang2023differentiable, karkus2023diffstack},};
        \node[anchor=north west] at (10.72,2.09){\cite{ huang2024dtpp}};
        
        \node[anchor=north west] at (4.87,1.28){\cite{hu2023planning, jiang2023vad, ye2023fusionad, chen2024vadv2, zheng2025genad}};

    \end{tikzpicture}
    \caption{General overview of architectural AD stack approaches aligned with the SA model \cite{endsley1995toward}.}
    \label{fig:Overview_AD_Stack}
\end{figure*}
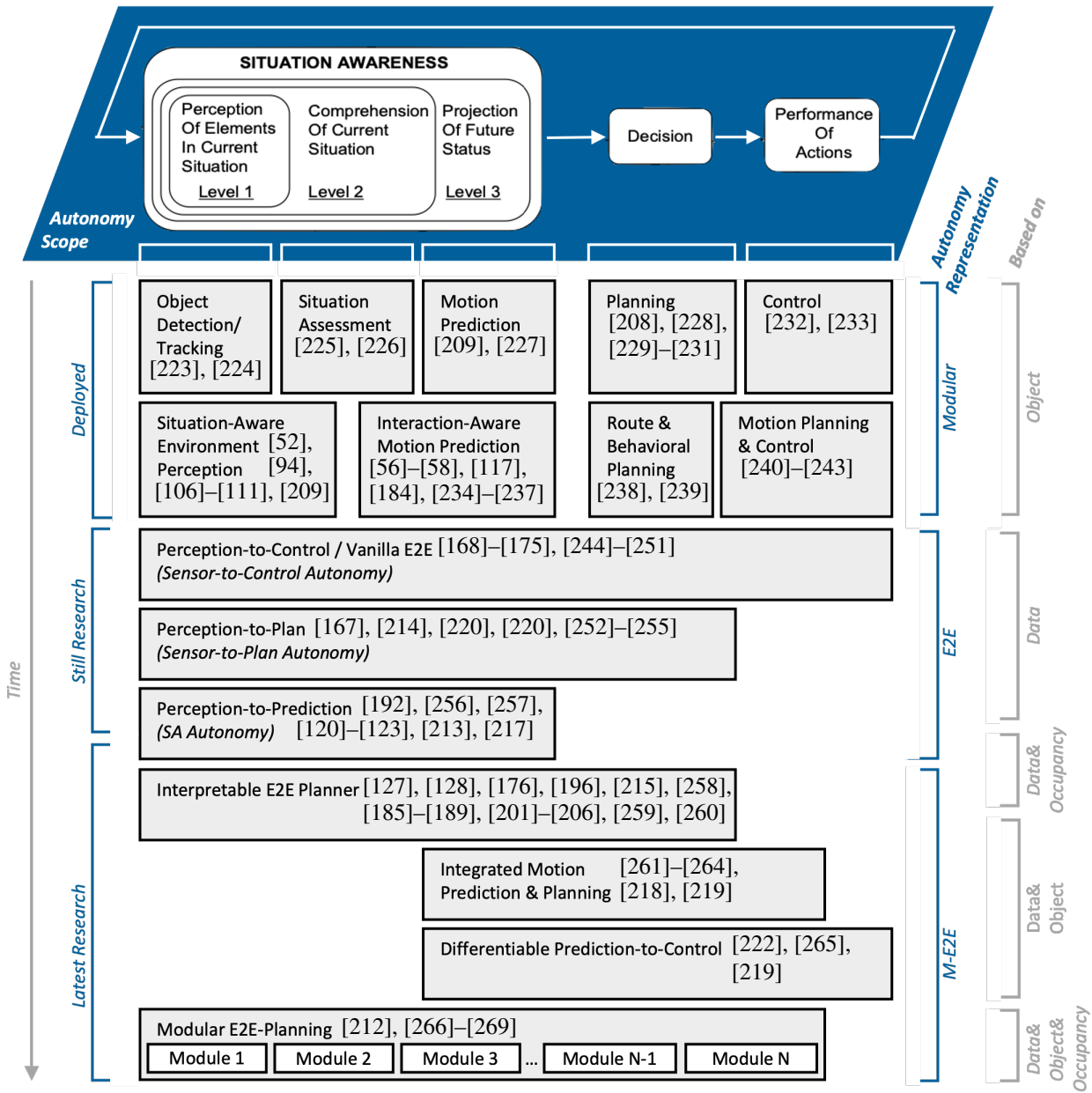

However, besides modular and E2E AD architectures various intermediate architectures are currently under investigation. For instance, there exist AI-driven E2E architectures that improve the interpretability, e.g., by means of intermediate auxiliary objectives or representations \cite{zeng2019end,zeng2020dsdnet, sadat2020perceive, casas2021mp3, cui2021lookout, chitta2021neat, chen2022learning, hu2022st, hu2022model}, or even through in-cooperated prior knowledge \cite{zeng2019end, philion2020lift, casas2020importance}. In this regard, FMs for AD, which provide text-based scene interpretation and decision-making explanation, are also considered as interpretable E2E AD systems \cite{cui2023drivellm, sima2023drivelm, xu2024drivegpt4, you2024v2x, tian2024drivevlm}.
In contrast, the modular-driven AD stacks increasingly integrate the subtasks of detection, tracking, prediction and planning, e.g., through joint optimization \cite{sadat2019jointly, casas2018intentnet, zhou2020tracking, casas2020implicit,  weng2022mtp, weng2022whose}, downstream metrics in-cooperation \cite{ivanovic2022injecting, hu2023planning}, or even joint task coverage \cite{liang2020pnpnet, gu2023vip3d, renz2022plant}. Since the prevailing object-driven detection, tracking, and prediction alongside SA is limited \cite{sadat2020perceive, biswas2024quad}, several researchers argue in favor of treating SA at once. In particular, by directly predicting temporal-spatial occupancy maps \cite{hoermann2018dynamic, mahjourian2022occupancy, liu2022strajnet, tong2023scene, liulet, li2025viewformer}. In this way, existing SA shortcomings such as limited information and uncertainty propagation or vulnerability to distribution shifts are addressed at the expense of SA interpretability. Furthermore, current research explores the growing intertwining of prediction and planning towards bidirectional frameworks \cite{huang2023gameformer, huang2024dtpp, jia2023driveadapter, hagedorn2024integration}, alongside the pursuit of differentiability across AI-driven and traditional systems, e.g. from prediction to control \cite{karkus2023diffstack}. A general overview of the various architectural approaches is presented in Figure \ref{fig:Overview_AD_Stack}, which is aligned with the SA model \cite{endsley1995toward}.

As shown in Figure \ref{fig:Overview_AD_Stack}, besides modular and vanilla E2E AD stacks various intermediate architectures are emerging to combine the advantages of both paradigms. Depending on their origin, these architectures are referred to as interpretable E2E architectures, modular E2E planning \cite{chen2023end}, differentiable and modular \cite{karkus2023diffstack} or planning-oriented AD stack \cite{hu2023planning}. Overall, current research generally emphasizes the need for a certain degree of modularity for interpretability combined with data-driven E2E adjustments and downstream task alignment, to achieve targeted outcomes. Throughout this paper, this set of AD stack architectures is referred to as modular E2E (M-E2E) architectures. These M-E2E AD stacks acknowledge and address limitations in AD and offer initial insights into measures required to progress toward fully autonomous driving. In this regard, the references in Figure \ref{fig:Overview_AD_Stack} demonstrate, that AI is used across all stages and architectural designs. Thereby, transformers are predominantly used in current approaches \cite{zeng2022motr, ngiamscene, shi2022motion, li2022bevformer, renz2022plant, hu2023planning}, while foundation models (FMs) emerge as an active and promising research area, offering potential for improved generalization yet presenting numerous open research questions \cite{zhou2024vision, sima2023drivelm, xu2024drivegpt4, you2024v2x}. Accordingly, while classical modular AD stack are still dominantly deployed, M-E2E concepts are also emerging in the industry, e.g. Tesla and Wayve \cite{chen2023end}. 

\subsection{Discussion}
It becomes apparent, that the classical modular AD stack is limited in multiple perspectives. The SA-driven analysis reveals, that the situation understanding, Level 2 SA, is particularly restricted. Increased interconnectivity to reduce complexity reinforces rethinking modularity towards increased flexibility. In addition, the semantic gap remains a significant challenge, underscoring the importance of dynamic decision-making and advanced cognitive processes. Although recent research has identified and started addressing several limitations, the realization of fully autonomous driving is still a long-term objective. Therefore, the subsequent section analyzes the steps required to advance autonomy.

\section{Towards Fully Autonomous Driving}\label{AutonomousDriving}

This section deals with the open question of how the challenges carved out in the previous section can be addressed and reconciled with current technological developments. Thereby, this section directly builds up on the latest research. To this end, possible further developments towards autonomous driving are identified.

\subsection{From Hard to Flexible Connections}

Based on the analysis of the current SOTA in Section \ref{AutomatedDriving}, achieving fully autonomous driving requires a more comprehensive transfer of self-responsibility and, above all, fewer hard-coded connections, e.g., between modules. Especially as hard-coded connections represent hand-crafted, knowledge-induced structures that limit the autonomy required to overcome open-world challenges. Therefore, in addition to the general shift in engineering processes toward iterative development and renewal \cite{ullrich2024expanding}, evolving the underlying AD stack to achieve greater flexibility is essential. However, not only the functional requirements but also the immense demands on data processing and computational efficiency emphasize the importance of moving towards more flexible, adaptable, and less prevalent AD stacks.

Based on current technological knowledge, one way to achieve human-like dynamic, context-specific flexibility, and adaptability for appropriate decision-making in complex situations could be the amplified use of attention mechanisms \cite{vaswani2017attention}. These mimic cognitive attention and provide soft weighting depending on the context. In other words, the weights and, thus, the interconnection of individual modules can be varied at runtime depending on the context. While this approach is used in individual sub-modules such as perception \cite{tian2020sa, wang2023sat}, prediction of other road users \cite{xie2020sast, shi2022motion, lin2020self}, or planning \cite{ye2021gsan, wen2022model, hu2023planning}, it is also conceivable to use it across the entire information processing chain in the area of interfaces. In this way, attention is used in LLMs \cite{ray2023chatgpt, kitaev2018multilingual}, which have already been applied in the AD context to tasks such as vision \cite{zhou2024vision} or prediction \cite{seff2023motionlm}.

An overview of the internal interfaces within the AD stack of currently emerging approaches is provided in Table \ref{tab:comparison_AI_AD_stack_interfaces}. While inspired by Figure \ref{fig:Overview_AD_Stack}, the table focuses on a promising subset of AD stack methods that achieve a balance between modularity and flexibility, while considering the increasing intertwining of functionalities. These approaches are particularly interpretable and modular E2E approaches as well as integrated and/or differentiable prediction to planning concepts.

As can be seen from Table \ref{tab:comparison_AI_AD_stack_interfaces}, the interpretable E2E (I-E2E) interfaces relied initially on semantic representations, which are also widely used in traditional modular AD stacks. More recently, however, latent, query-based and, in particular, token-based interfaces have also been introduced. While latent representations predominate in vanilla E2E approaches, they are less prevalent in the area of I-E2E methods, as the implicit characteristic has a negative impact on interpretability. Moreover, there are also token-based ex-/implicit interfaces used within I-E2E AD stacks. Depending on their extraction from semantic or latent variables, these tend to be implicit or explicit. Through their use within the I-E2E approaches by means of FMs, such as LLMs, or VLMs , they offer increased situation assessment due to their generalization and interpretation capability, regardless of the internal interface characteristics. 

\begin{table*}[h]
	\centering
	\caption{Comparison of AD stack internal interfaces of interpretable E2E (I-E2E), integrated motion prediction and planning (I-M-P\&P), differentiable prediction-to-control (DIFF) and modular E2E planning (M-E2E-P) approaches considered in Fig.\ref{fig:Overview_AD_Stack}.}
		\label{tab:comparison_AI_AD_stack_interfaces}
		\begin{tabular}{l l l l l l }
			\toprule 
            & \textbf{Method} & {\textbf{Interface}} &  {\textbf{Characteristic}} & {\textbf{Details}}  \\
            \midrule
             \multirow{16}{0cm}{\rotatebox{90}{I-E2E}} &NMP'19 \cite{zeng2019end} & semantic rep. & explicit &  3D detections, future motion forecast, space-time cost volume   \\
            & P3'20 \cite{sadat2020perceive} &semantic rep.  & explicit & lidar \& map features, occupancy rep., recurrent occupancy forecasting  \\
            &DSDNet'20 \cite{zeng2020dsdnet}  & semantic rep. & explicit & feature map, actor \& trajectory feature, 2D BEV waypoints   \\
            & LSS'20 \cite{philion2020lift} & semantic/latent rep. & mixed & BEV scene \& cost map, segmentation, frustums, latent depth distrib.\\
            & LookOut'21 \cite{cui2021lookout} & latent rep. & implicit & implicit latent variable model, latent sampler \\
            & MP3'21 \cite{casas2021mp3} & semantic rep. & explicit & geometric \& semantic features, online map, dynamic occupancy field \\
            & NEAT'21 \cite{chitta2021neat} & semantic/latent rep. & mixed & image features, attention map, NEAT feature, BEV semantic pred.  \\
            & LAV'22 \cite{chen2022learning} & semantic/latent rep.  & mixed & spatial features, RoI features, sparse pillar features, embeddings \\
            & ST-P3'22 \cite{hu2022st} & semantic rep. & explicit & BEV rep. and future scenes predictions \\
            & MILE'22 \cite{hu2022model} & latent variables & implicit & temporal dynamics, deterministic history \\
            & DriveLM'23 \cite{sima2023drivelm} & token-based & ex-/implicit & graph-structured question answering \\
            & DriveLLM'23 \cite{cui2023drivellm} & semantic rep. & explicit &bidirectional semantic LLM AD stack bridge\\
            & Drive-GPT4'24 \cite{xu2024drivegpt4} & token-based & ex-/implicit & text and video tokens \\
            & DriveVLM'24 \cite{tian2024drivevlm} & token-based & ex-/implicit & image- and text-tokens \\
            & V2X-VLM'24 \cite{you2024v2x} & token-based & ex-/implicit & textual prompts and multimodal visual inputs\\
            & QuAD'24 \cite{biswas2024quad} & query-based & hybrid & spatio-temporal occupancy query points \\
            & Alphamayo-R1'25  \cite{wang2025alpamayo} & token-based & ex-/implicit & 
           textual reasoning trace  \\
 			\midrule
            \multirow{5}{0cm}{\rotatebox{90}{I-M-P\&P}}&  PiP'20 \cite{song2020pip} & latent rep. & implicit &  planning-, observation-, target, fused-target-tensor \\
            & FutureFree'21 \cite{hu2021safe} & semantic rep. & explicit & freespace-centric rep. from sensors \& forecasted future freespace \\ 
            &  SafePathNet'23 \cite{pini2023safe} & query-based/(semantic rep.) & hybrid & SDV- \& agent-queries internal, (semantic rep. in- \& output) \\
            &  IJP'23 \cite{chen2023interactive} & semantic rep. & explicit & reference-, ego-sample-, predicted-trajectories \\
            &  GameFormer'23 \cite{huang2023gameformer} & query-based & hybrid & agent history-, modality embedding-, future-query in single transformer   \\
            \dashmidrule  
            &DTPP'24\cite{huang2024dtpp}  & semantic rep. & explicit  &  tree-structured planner, trajectory- \& scenario-tree \\
            \dashmidrule
            \multirow{2}{0cm}{\rotatebox{90}{DIFF}}& DIPP'23 \cite{huang2023differentiable} & semantic rep./query-based & hybrid  & multi-agent predictions, ego-trajectory, agent-query  \\
            &DiffStack'23\cite{karkus2023diffstack} & semantic rep. & explicit &  multi-modal trajectory predictions, ego trajectories \\
            \midrule
			\multirow{5}{0cm}{\rotatebox{90}{M-E2E-P}}& UniAD'23 \cite{hu2023planning} &query-based & hybrid & map-, track-, motion-, occ.-queries   \\ 
            & VAD'23 \cite{jiang2023vad} & query-based & hybrid & map-, BEV-, agent-, ego-queries   \\ 
            & FusionAD'23 \cite{ye2023fusionad} & query-based & hybrid & BEV, plan, ego, motion-queries   \\ 
            & VADv2'24 \cite{chen2024vadv2} & token-based & ex-/implicit & map-, image-, agent-, traffic, planning-tokens   \\ 
            & GenAD'24 \cite{zheng2025genad}  & query-/token-based & mixed & map-, BEV-, agent-, ego-tokens, self-, deformable-/cross-attention queries  \\ 
			\bottomrule \\[-6pt]
	\end{tabular}
\end{table*}

In comparison, the approaches of integrated motion prediction and planning (I-M-P\&P) and differentiable prediction-to-control (DIFF) rely mainly on semantic and query-based interfaces. In addition, query-based interfaces dominate in the area of modular E2E planning (M-E2E-P). Query-based interfaces are fundamentally built on attention mechanisms and are classified as hybrid, as they usually use explicit queries to extract implicit information. While almost all of the listed approaches in Table \ref{tab:comparison_AI_AD_stack_interfaces} use transformers and attention mechanisms internally for specific sub-functionalities, this shows that the use of attention mechanisms is also increasing across sub-functionalities in terms of interfaces. The combination with token-based representations, which can be used as precisely defined, discretized and modularized internal information bricks, as has been done in e.g. \cite{zheng2025genad}, appears promising. 

However, the pure use of FMs and token-based interfaces along AD stacks does not lead to the desired intertwining of the different sub-modules or -functionalities. Although FMs are currently popular and achieve considerable performance in SA, they do not enable reasoning in the true sense, whereas approaches such as those in M-E2E-P with query-based interfaces that build up on various attention mechanisms appear more promising. Moreover, query-based interfaces uniquely enable flexibility through request-driven mechanisms, allowing for adaptive and self-responsible information flow guidance throughout data processing and information compression. 

Accordingly, with respect to the AD stack, attention mechanisms connecting various modules could be interpreted as an open long-tail distribution countermeasure at the highest AD system level. This means that attention mechanisms addressing the issue of handcrafted, rule-based AD stack architectures at the highest system level and could be considered as learning-based, self-responsible system orchestrators. Nevertheless, other options are also conceivable, e.g., orchestrators with increased flexibility \cite{kampmann2019dynamic} in combination with a differentiable and modular AD stack \cite{karkus2023diffstack}. However, finding a suitable cost function that is needed in such orchestrators could be challenging. For instance, this is evident in the subarea of trajectory planning, which is why learned weights or even entire learned cost functions are increasingly being used \cite{hagedorn2024integration}. Although learned cost functions reduce knowledge induction, they are usually not context-specific. However, it stands to reason that in an optimization-based orchestration of services, the costs and weightings of individual aspects such as safety and progress could also be adjusted in a context-specific manner. In comparison, attention mechanisms enable such context-specific adjustments implicitly and also have no specific requirements for individual modules as they are able process implicit and explicit information. Nevertheless, attention mechanisms have also respective demands, e.g., they implicitly require within the given application cross-context training of the AD stack. It turns out that there are various possibilities foreseeable with individual pros and cons. Nevertheless, attention mechanisms have largely proven their capabilities. Furthermore, they are implicitly appropriate for highly autonomous orchestration. Therefore, query-based, attention-driven interfaces and orchestration could be quite promising. 

\subsection{Towards Improved Situation Awareness}

As outlined in Section \ref{AutomatedDriving}, SA is key towards fully autonomous driving. A concise and nuanced scene, situation, scenario, and context representation, i.e., SA in general, is crucial from a functional perspective.  In addition, the individual SA level representations are also relevant for system interpretability. To review the state of research in this regard, Table \ref{tab:comparison_AI_method_SA} provides a more detailed overview of the different levels of SA. In addition to the I-E2E, I-M-P\&P, DIFF and \text{M-E2E-P} of Table \ref{tab:comparison_AI_AD_stack_interfaces}, the overview in Table \ref{tab:comparison_AI_method_SA} also considers SA, Sensor-2-Plan (S2P) and vanilla E2E autonomy approaches. In particular, the various black-box-like autonomy methods are also considered here, as a consistent, constantly updated context consideration is desirable, which could be accounted for in both interpretable and non-interpretable intertwined AD stack approaches.

\begin{table*}
	\centering
	\caption{Overview of the SA consideration and representation across various AD stack levels and concepts.}
	\label{tab:comparison_AI_method_SA}
	\resizebox{\textwidth}{!}{
	\begin{tabular}{l l l l l }
			\toprule
			& \textbf{Method}  & {\textbf{Level 1 SA}} & {\textbf{Level 2 SA}} & {\textbf{Level 3 SA}}\\
            \midrule
            \multirow{16}{0cm}{\rotatebox{90}{Vanilla E2E Autonomy}} & E2E-CNN'16\cite{bojarski2016end}  & \xmark/CNN & \xmark/CNN & \xmark/CNN\\
            & DRL-Framework'17\cite{sallab2017deep}  & latent input vector& spatial features & \xmark/RL \\
            & CIL'18 \cite{codevilla2018end}  & joint input representation & \xmark/CIL & \xmark/CIL \\
            & CILRS'19 \cite{codevilla2019exploring} & latent perception state & \xmark/CIL & \xmark/CIL \\ 
            & UD-CIL'20\cite{hawke2020urban}  & intermediate learned perception features & \xmark/CIL & \xmark/CIL \\
            & MaRLn'20 \cite{toromanoff2020end}  & \xmark/RL & \xmark/RL & predict affordances/RL state \\
            & DeepImitative'20\cite{Rhinehart2020Deep} & latent (MapFeat, PastRNN) & latent (MapFeat, JointFeat) & latent (FutureMLP, FutureRNN)\\ 
            & LSD'20 \cite{ohn2020learning}  & \xmark/BC & context-embeddings/BC & situation-specific policy predictions/BC \\
            & LBC'20 \cite{chen2020learning}  & \xmark/teacher-guided policy learning & \xmark/teacher-guided policy learning  &  \xmark/teacher-guided policy learning  \\
            & WOR'21 \cite{chen2021learning} & \xmark/learned policy & \xmark/learned policy & \xmark/learned policy\\
            & Roach'21 \cite{zhang2021end} & BEV rep. \& measurement vector & \xmark/IL-agent\&RL-coach & \xmark/IL-agent\&RL-coach \\
            & TCP'22 \cite{wu2022trajectory} & image \& measurement features & \xmark & trajectory-guided control prediction  \\
            & UrbanDriver'22\cite{scheel2022urban} & vectorized rep. & mid-level vectorized rep.& differentiable traffic simulator \\
            & TRAVL'22 \cite{zhang2022rethinking} & BEV tensor & \xmark/learned policy& imagined data \& look-ahead \\
            & ThinkTwice'23 \cite{jia2023think} & BEV features & compact env. \& mission vector & prediction feature \\
            & Hydra-MDP++'24 \cite{lihydra} & image \& lidar tokens & environment token & \xmark/teacher-student model\\
            \midrule
            \multirow{6}{0cm}{\rotatebox{90}{S2P Autonomy}} & ALVINN'88 \cite{pomerleau1988alvinn} & \xmark/NN  & \xmark/NN  & \xmark/NN \\
            & Seg2WP'18 \cite{mueller18a} & segmentation map & \xmark /CIL & \xmark /CIL \\
            & TransFuser'22 \cite{chitta2022transfuser} & auxiliary map, depth, bounding boxes & auxiliary semantic seg. &
            feature vector of global context \\
            & PlanT'22 \cite{renz2022plant}  & input \& embedding tokens & learnable [CLS] tokens  & auxiliary vehicle future prediction \\
            & DriveAdapter'23\cite{jia2023driveadapter}  & BEV features & \xmark/teacher-student model & \xmark/teacher-student model \\
            & DriveMLM'23 \cite{wang2023drivemlm} & image \& cloud token embeddings& (\xmark) & LLM-based reasoning \& explanation  \\
            \midrule
            \multirow{6}{0cm}{\rotatebox{90}{SA Autonomy}} 
            & EmergentOccupancy'22\cite{khurana2022differentiable} & ego-centric freespace &  (\xmark) & future occupancy predictions \\
            & OccFlow'22\cite{mahjourian2022occupancy} & BEV map & occupancy grids & occupancy flow flields \\
            & OccNet'23 \cite{tong2023scene} & BEV feature & semantic scene completion & (flow annotation) \\
            & ImplicitO'23 \cite{agro2023implicit} & feature map & (\xmark) & occupancy probability \& flow over time \\
            & LetOccFlow'24 \cite{liulet} & BEV feature & 3D occupancy & 3D occupancy flow \\
            & ViewFormer'25 \cite{li2025viewformer} & BEV feature & 3D occupancy & occupancy flow \\
            \midrule
            \multirow{16}{0cm}{\rotatebox{90}{I-E2E}} & NMP'19 \cite{zeng2019end} &  3D detections  & (\xmark) & motion forecast \\
            &P3'20 \cite{sadat2020perceive} &  lidar \& map features & fused features & semantic occupancy forecast\\
            &DSDNet'20 \cite{zeng2020dsdnet} & feature map \& detection & (\xmark) & probab. multimodal social pred. \\
            &LSS'20 \cite{philion2020lift} & BEV rep. & BEV semantic segmentation & BEV predictions \\
            &LookOut'21 \cite{cui2021lookout} & object detector & actor context & latent scene dynamics \& decoder \\
            &MP3'21 \cite{casas2021mp3} & geometric \& semantic features & online map & dynamic occupancy field \\
            &NEAT'21 \cite{chitta2021neat} & 2D image features & neural attention fields & BEV semantic prediction \\
            &LAV'22 \cite{chen2022learning}  & map-view features & (\xmark) & multi-modal trajectory predictions \\
            &ST-P3'22 \cite{hu2022st} & BEV features & (\xmark) & dual pathway \& occupancy fields \\
            &MILE'22 \cite{hu2022model} & BEV features & latent rep. world model & latent state trajectory forecasting  \\
            &DriveLM'23 \cite{sima2023drivelm} & graph visual question answering (GVQA) & GVQA-based reasoning & GVQA-based prediction \\
            &DriveLLM'23 \cite{cui2023drivellm} & any existing AD stack & LLM-based reasoning & LLM-based decision-making \\
            &Drive-GPT4'24 \cite{xu2024drivegpt4} & question-answering (QA) & QA-based reasoning & QA-based prediction \\
            &DriveVLM'24 \cite{tian2024drivevlm}  & chain-of-though (CoT) description& CoT-based analysis & CoT-based prediction \\
            &V2X-VLM'24 \cite{you2024v2x} & scene description prompting & VLM-based interpretation& (\xmark) / VLM internal motion prediction\\
            &QuAD'24 \cite{biswas2024quad} & BEV latent rep. & BEV \& occupancy & implicit occupancy model\\
            & Alphamayo-R1'25  \cite{wang2025alpamayo} & multi-camera cideo tokens & chain-of-causation (CoC) & Cosmos-Reason \cite{azzolini2025cosmos} based prediction \\
            
%
%
            \midrule
            \multirow{6}{0cm}{\rotatebox{90}{I-M-P\&P}}&  PiP'20 \cite{song2020pip} & observation tensor & social context encoding & fused target encoding  \\
            & FutureFree'21 \cite{hu2021safe} & freespace & (\xmark)  & forecasted future freespace  \\
            &  SafePathNet'23 \cite{pini2023safe} & \xmark & (assume: vectorized scene rep.)  &  SDV \& agent future trajectories \\
            &  IJP'23 \cite{chen2023interactive} & \xmark & \xmark  & predicted trajectories  \\
            &  GameFormer'23 \cite{huang2023gameformer} &  \xmark & \xmark & game theory inspired transformer \\
            \dashmidrule  
            &DTPP'24\cite{huang2024dtpp} & \xmark & env. encoding & ego-conditioned predictions  \\
            \dashmidrule
            \multirow{2}{0cm}{\rotatebox{90}{DIFF}} & DIPP'23 \cite{huang2023differentiable} & \xmark & local scene context encoder & future decoder \\
            &DiffStack'23\cite{karkus2023diffstack} & \xmark & \xmark & Trajectron++ \cite{salzmann2020trajectron} \\
            \midrule
            \multirow{5}{0cm}{\rotatebox{90}{M-E2E-P}} &
			UniAD'23 \cite{hu2023planning} & BEV features & (\xmark)  /  MotionFormer  & OccFormer \\
            & VAD'23 \cite{jiang2023vad} & BEV features & vectorized map & vectorized agent motion \\
            & FusionAD'23 \cite{ye2023fusionad} & BEV camera, lidar features & fused BEV feature & modality self-attn. \& refinement net.  \\
            & VADv2'24 \cite{chen2024vadv2} & (\xmark) / map \& image tokens & scene tokens & agent \& traffic element tokens \\
            & GenAD'24 \cite{zheng2025genad} & BEV tokens & scene tokens & latent future generation   \\
			\bottomrule \\[-6pt]
	\end{tabular}}
\end{table*}

As shown in Table \ref{tab:comparison_AI_method_SA}, both vanilla and S2P autonomy approaches often rely on a holistic approach to learning policies, i.e., mappings between inputs (e.g., sensors) and outputs (e.g., trajectory plans or control inputs). While early approaches considered neural networks (NN) or convolutional neural networks (CNN), there was first research on reinforcement learning (RL), behavioral cloning (BC), and imitation learning (IL), whilst recently developed approaches concentrate on conditional imitation learning (CIL) and teacher-student models. All of these approaches implicitly take SA into account and do not model the scene or context considerations. In contrast, within SA autonomy, the various SA levels are considered more explicitly from BEV over occupancy map to occupancy (flow) predictions/field. Thus, providing inspiration for a continuous context representation for improved SA. In contrast to the traditional modular AD stack, SA Autonomy follows an occupancy-based rather than an object-based approach, which expands the scope of responsibility of the used AI method as there is less hand-crafted object-based reasoning. On the one hand, occupancy-based approaches show good performance and improved generalization abilities, on the other hand, human reasoning is object-driven.

Within the I-E2E subset in Table \ref{tab:comparison_AI_method_SA}, two main directions exist. On the one hand, some approaches consider Level 2 SA only implicitly and to a limited extent, as indicated by (\xmark) in the Level 2 SA column. On the other hand, some FM-based approaches address Level 2 SA constraints in particular but have inherent FM-related concerns and restrictions. Apart from that, the I-M-P\&P and DIFF methods mainly deal with Level 3 SA but indicate the relevance of scene/environment encoding for subsequent forecasting.

Furthermore, M-E2E-P within Table \ref{tab:comparison_AI_method_SA} considers all SA levels successively and continuously like SA autonomy. However, while within SA autonomy occupancy-based representations are dominant across Level 2 and 3 SA, within M-E2E-P there is a greater conceptual diversity. Reaching from vectorized and token-based Level 2 SA representations to a multitude of different Level 3 SA representation concepts. In general M-E2E-P concepts bridge machine readability with human interetability along SA. Thereby, some approaches tend to maintain machine-readable representations internally along processing, while branching human-readable conversions out for interpretability. Others focus on machine- and human-readable interfaces between functionalities.

Overall, it becomes apparent, that diverse concepts to consider SA relevant context information exist. While vanilla E2E and S2P autonomy approaches predominantly focus on implicit handling, I-E2E, I-M-P\&P, DIFF, and M-E2E-P target more explicit scene and context modeling. Thereby, not all representations are directly human-interpretable. Moreover, the capabilities of FMs such as LLMs, VLMs, and VLAs are promising to supplement existing AD stacks. For instance, \cite{wang2023drivemlm} enhances existing AD stacks using a multi-modal LLM (MLM). Nevertheless, while FMs are promising and showing impressive results, they typically do not provide the human-like reasoning, that is necessary for fully autonomous driving. Yet, latest research demonstrates via Nvidia’s Alpamayo-R1 \cite{wang2025alpamayo}, a VLA model that integrates Chain of Causation (CoC) reasoning on top of the physically pretrained VLM backbone Cosmos-Reason1 \cite{azzolini2025cosmos}, thereby providing a first step toward improved reasonability. Despite this, we imagine FM as eighter one functional capability block or parallel integrated system. For instance, as one functional capability and suitable human-machine interface (HMI), that provides situation interpretations in tokenized representations and textual descriptions. Or parallel integrated to classical methods \cite{Bosch, yao2025navigating} as System 1 and System 2 framework in accordance to \cite{kahneman2011thinking}. In general, in combination with query-based attention interfaces information flow of various representations and sources could be promising. 

\begin{figure}[]
	\centering	
	\includegraphics[width=\linewidth]{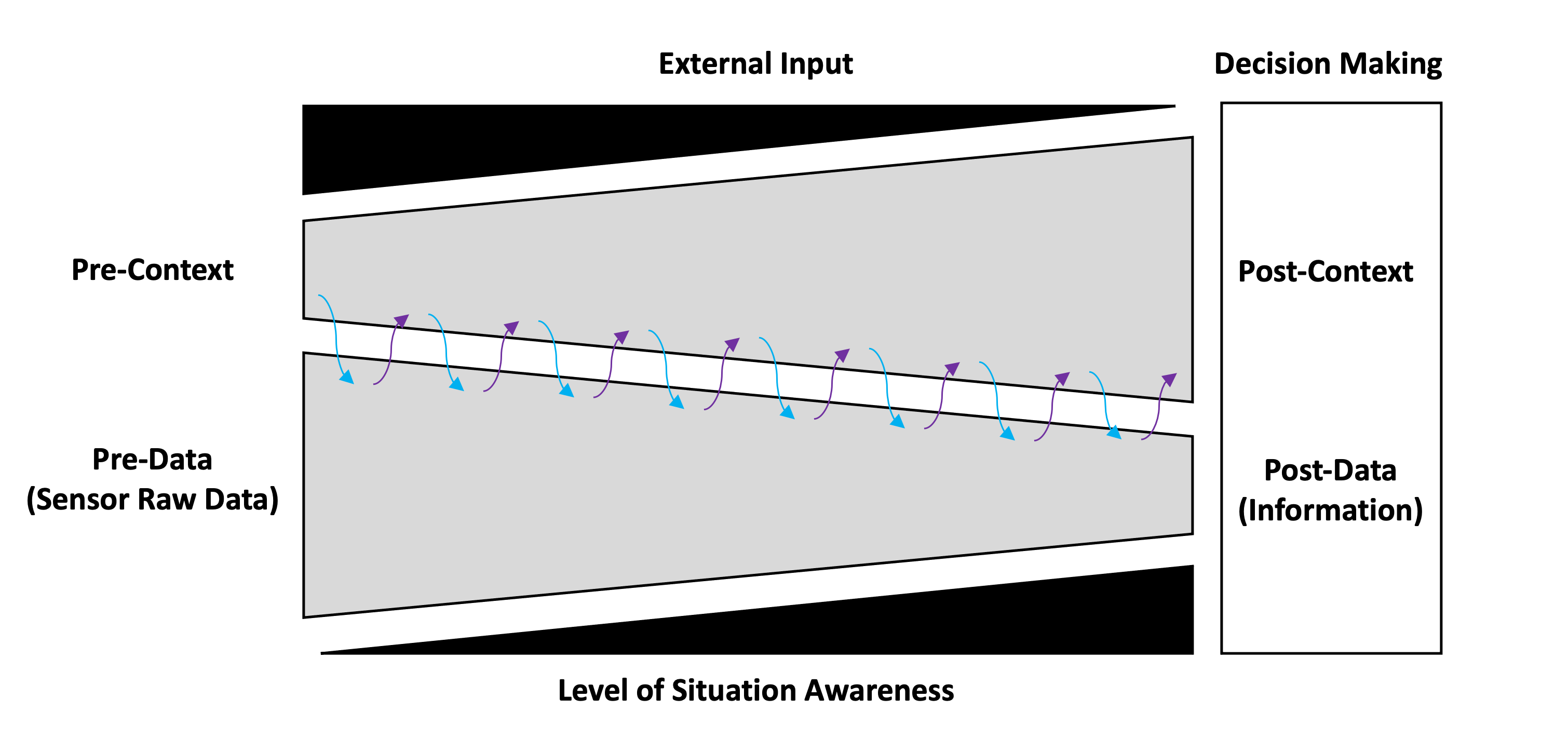}
	\caption{Visualization of the directed context consideration. At the beginning of the processing chain there is a lot of data as well as external information. With increasing processing depth, the data can be sorted out and consolidated in a highly aggregated manner, which results in a context gain in parallel.}
	\label{fig:contextdata}
\end{figure}

One conceptual way of improving SA, in general, is a directed information processing chain across the entire AD stack in which all data is constantly acquired and processed to generate a consistent, constantly updated context representation. Thereby, e.g., the initial context knowledge can be used to reduce the amount of data in the individual processing steps. New context knowledge can be derived from the processing steps. Thus, with increasing processing depth, the data volume can be reduced while context knowledge can be enhanced simultaneously to provide highly aggregated information. This way, data is only reduced and sorted out if it will no longer be necessary for downstream steps due to the context knowledge at hand. A respective conceptual approach is illustrated in Figure \ref{fig:contextdata}. 

Although the illustration is simplified, e.g., with respect to the overall actualization process, that is indispensable, as the context rapidly becomes outdated due to temporal and spatial changes in dynamic environments, the illustration suggests architectural implications. For instance, a context layer could be incorporated over the entire information processing chain from perception to action, which consistently summarizes the increasing knowledge about the context. This context status could in turn act on the interfaces along the AD stack and functionalities in general. While such an approach seems reasonable, it raises the open question of how the context should be modeled. Advances in the field of AI, especially in context-based AI systems, indicate that the use of latent representations and tokenization appears to be effective. Nevertheless, while this is evidenced by the progress made by AI systems, on the other hand, visual scene representations are used as context information \cite{casas2021mp3, sadat2020perceive} to increase interpretability. However, these visual representations are usually limited to a few contextual aspects. A different direction is demonstrated by latest FM, that enable tokenized context aggregation along chain of thought (CoT) or chain or chain of causation (CoC) and textual explenations for interpretations \cite{tian2024drivevlm, wang2025alpamayo}. Overall, whether a latent representation, tokenized representation or a combination with a simplified situation representation is ultimately the most effective way is an area of potential further research.  

Finally, it should be noted, that the analysis of methods in Table \ref{tab:comparison_AI_method_SA} shows, that first steps towards such a context consideration are made. For instance, through BEV information compression and successive reuse. Nevertheless, current efforts are limited in several respects. Especially, as either or is currently predominant, as described in more detail in the following. Currently, one sensor source and one context representation are mostly employed. For instance, either occupancy-based or object-based, either more traditional or FM-based, either LiDAR-based or vision-based is applied. Although this reduces the complexity in the already wide-ranging AD stack research, it also limits the capabilities. In addition, the step-by-step consideration of the context along the processing chain is predominant. However, recent research results show that functionalities along the AD stack must also be intertwined across successive processing steps. Therefore, a dedicated and decoupled context layer that evolves over space and time appears to be promising. 

\subsection{External Sourcing of Contexts} 

In addition to internal context aggregation and interpretation, querying external context is a valuable complement to on-board capabilities. This involves connected and automated vehicles (CAVs) using vehicle-to-anything (V2X) communication, standardized for example by the European Telecommunications Standards Institute (ETSI) \cite{etsiintelligent663}. Key message formats include cooperative awareness messages (CAM) \cite{etsiintelligent637}, collective perception messages (CPM) \cite{etsiintelligent562}, and maneuver coordination messages (MCM) \cite{etsiintelligent578}, which facilitate the exchange of contextual information to enhance external sensing and situational awareness. The Infrastructure Support Levels for Automated Driving (ISAD) \cite{inframix2021isadlevels} provides a classification scheme for road infrastructure capabilities, similar to Society of Automotive Engineers (SAE) levels for vehicle capabilities. Consistent with previous subsections, Table \ref{tab:comparison_AI_method_inputs} summarizes both internal and external sources utilized in the surveyed methods. It also highlights a subset of real-world commercial systems \cite{uber2020safetyreport, lyft2020safety, ArgiAI2021safety, Motional2021safety, waymo2021safetyreport, zoox2021safety, cruise2022safety, nvidia2024safety, mobileye2024safety, Tesla2021, Toyota2022, Mercedes2023} and V2X research efforts \cite{buchholz2021handling, cress2023intelligent, shan2024experimental, lampe2020reducing}. In general, it becomes apparent that while advanced digital infrastructure (ISAD levels C, B, and A) is increasingly explored in V2X research, advancing AD stack architectures and existing commercial systems predominantly operate under conventional infrastructure conditions (ISAD levels D and E).

\begin{table*}
	\centering
	\caption{Comparison of internal and external sources used in the methods surveyed.}
	\resizebox{\textwidth}{!}{
		\label{tab:comparison_AI_method_inputs}
		\begin{tabular}{l l l l l }
			\toprule
			& \textbf{Method}  & {\textbf{Internal Sources}} & {\textbf{External Sources}} & {\textbf{Classification}}\\
            \midrule
            \multirow{16}{0cm}{\rotatebox{90}{Vanilla E2E Autonomy}} & E2E-CNN'16\cite{bojarski2016end}  & multi-view images & \xmark & vision-based \\
            & DLR-Framework'17\cite{sallab2017deep}  & raw sensor input vector & \xmark & n/a \\
            & CIL'18 \cite{codevilla2018end}  & camera images, ego speed, high-level command & \xmark & mapless, vision-driven \\ 
            & CILRS'19 \cite{codevilla2019exploring} & single image, ego speed, high-level command & \xmark & mapless, vision-driven\\ 
            & UD-CIL'20\cite{hawke2020urban}  & multi-view images, route command & \xmark & mapless, vision-based \\
            & MaRLn'20 \cite{toromanoff2020end}  & camera images & \xmark & mapless, vision-based\\
            & DeepImitative'20\cite{Rhinehart2020Deep} & mainly Lidar (possible camera images or both) & traffic light signal & mapless\\
            & LSD'20 \cite{ohn2020learning}  & camera images, ego speed, high-level command & \xmark & mapless, camera-driven \\
            & LBC'20 \cite{chen2020learning}  & camera images, ego speed, high-level command & (map to train privileged agent) & mapless, camera-driven \\
            & WOR'21 \cite{chen2021learning} & camera images, ego speed, high-level command & (maps to train world model) & mapless, camera-driven \\
            & Roach'21 \cite{zhang2021end} & measurement \& (BEV images) vector & \xmark & intermediate rep. \\
            & TCP'22 \cite{wu2022trajectory} & camera images, ego speed, high-level command & \xmark & mapless, vision-driven\\
            & UrbanDriver'22\cite{scheel2022urban} & (perception module output) & HD map & intermediate rep.  \\
            & TRAVL'22 \cite{zhang2022rethinking} & (agents motion history) & HD map & intermediate rep.  \\
            & ThinkTwice'23 \cite{jia2023think} & consecutive LiDAR sweeps \& RGB cameras & \xmark & mapless, multi-modal \\
            & Hydra-MDP++'24 \cite{li2024hydra} & camera images, measurements, high-level command & \xmark & mapless, vision-driven \\
            \midrule
            \multirow{6}{0cm}{\rotatebox{90}{S2P Autonomy}} & ALVINN'88 \cite{pomerleau1988alvinn} & camera images \& laser range finder & \xmark & mapless, multi-modal \\
            & Seg2WP'18 \cite{mueller18a} & camera images, high-level command & \xmark & mapless, vision-driven \\
            & TransFuser'22 \cite{chitta2022transfuser}  & LiDAR sweeps \& RGB cameras & \xmark & mapless, multi-modal  \\
            & PlanT'22 \cite{renz2022plant}  & object-based rep. build on \cite{zhou2020tracking}  & \xmark & intermediate rep. \\
            & DriveAdapter'23\cite{jia2023driveadapter}  & consecutive LiDAR sweeps \& multi-view images & (ground-truth BEV for teacher) & mapless, multi-modal \\
            & DriveMLM'23 \cite{wang2023drivemlm} & (textual AD bridge: LiDAR, camera, human input)  & (HD map API) & AD LLM enhancement \\
            \midrule
            \multirow{6}{0cm}{\rotatebox{90}{SA Autonomy}} 
            & EmergentOccupancy'22\cite{khurana2022differentiable} & consecutive LiDAR sweeps &  \xmark & mapless, LiDAR-based \\
            & OccFlow'22\cite{mahjourian2022occupancy} & past agent states & traffic light status, road structure & sparse agent \& env. states \\
            & OccNet'23 \cite{tong2023scene} & multi-view camera images & \xmark & mapless, vision-based \\
            & ImplicitO'23 \cite{agro2023implicit} & consecutive LiDAR sweeps & HD map & map- \& LiDAR-based\\
            & LetOccFlow'24 \cite{liulet} & temp. seq. multi-view camera images & \xmark & mapless, vision-based \\
            & ViewFormer'25 \cite{li2025viewformer} & multi-view camera images & \xmark & mapless, vision-based \\
            \midrule
            \multirow{16}{0cm}{\rotatebox{90}{I-E2E}} & NMP'19 \cite{zeng2019end} & consecutive LiDAR sweeps & HD map & map- \& LiDAR-based \\
            &P3'20 \cite{sadat2020perceive} & consecutive LiDAR sweeps & HD map & map- \& LiDAR-based \\
            &DSDNet'20 \cite{zeng2020dsdnet} & consecutive LiDAR sweeps & HD map & map- \& LiDAR-based  \\
            &LSS'20 \cite{philion2020lift} & $n$ images, incl. extrinsic \& intrinsic parameters & \xmark & mapless, vision-based \\
            &LookOut'21 \cite{cui2021lookout} & consecutive LiDAR sweeps & HD map & map- \& LiDAR-based  \\
            &MP3'21 \cite{casas2021mp3} & consecutive LiDAR sweeps, high-level command & \xmark & mapless, LiDAR-based \\
            &NEAT'21 \cite{chitta2021neat} & RGB cameras & \xmark & mapless, vision-based \\
            &LAV'22 \cite{chen2022learning} & consecutive LiDAR sweeps \& RGB cameras & \xmark & mapless, multi-modal \\
            &ST-P3'22 \cite{hu2022st} & multi-view images, high-level command & \xmark & mapless, vision-based \\
            &MILE'22 \cite{hu2022model} & camera images & \xmark & mapless, vision-only \\
            &DriveLM'23 \cite{sima2023drivelm} & front-view image & \xmark & mapless, LLM-based \\
            &DriveLLM'23 \cite{cui2023drivellm} & text & \xmark & mapless, LLM AD enhancement \\
            &Drive-GPT4'24 \cite{xu2024drivegpt4} & front-view RGB camera video sequence \& text& \xmark & mapless, multi-modal LLM \\
            &DriveVLM'24 \cite{tian2024drivevlm} & sequence of images & \xmark & mapless, VLM-based \\
            &V2X-VLM'24 \cite{you2024v2x} & images \& text & infrastructure images \& text & mapless, multi-modal LLM \\
            &QuAD'24 \cite{biswas2024quad} & consecutive LiDAR sweeps & HD map & map- \& LiDAR-based \\
            & Alphamayo-R1'25  \cite{wang2025alpamayo} & consecutive multi-view images, textual instructions & \xmark  & mapless, VLA incl. CoC \\
			\midrule
            \multirow{5}{0cm}{\rotatebox{90}{M-E2E-P}} &
			UniAD'23 \cite{hu2023planning} & multi-view images & \xmark & mapless vision-based \\
            & VAD'23 \cite{jiang2023vad} & multi-view images & \xmark & mapless, vision-based \\
            & FusionAD'23 \cite{ye2023fusionad} & consecutive LiDAR sweeps \& multi-view images & \xmark & mapless, multi-modal \\
            & VADv2'24 \cite{chen2024vadv2} & multi-view images & (map during training) & mapless in deployment \\
            & GenAD'24 \cite{zheng2025genad} & multi-view images & map information & map- \& vision-based \\
            \midrule
            \multirow{12}{0cm}{\rotatebox{90}{Real Sys. / Commercial}} &Uber ATG'18 \cite{uber2020safetyreport} & LiDAR, radar, camera, etc. & HD map & map-based, multi-modal \\   
            &Lyft'20 \cite{lyft2020safety} & LiDAR, radar, camera, etc. & HD map & map-based, multi-modal \\
            & Argo AI'21 \cite{ArgiAI2021safety} & LiDAR, radar, camera, etc. & HD map & map-based, multi-modal \\
            &Motional'21\cite{Motional2021safety} & LiDAR, radar, camera, etc. & HD map & map-based, multi-modal \\
            &Waymo'21 \cite{waymo2021safetyreport} & LiDAR, radar, camera, etc. & HD map & map-based, multi-modal \\
            & Zoox'21\cite{zoox2021safety} & LiDAR, radar, camera, etc. & HD map & map-based, multi-modal \\
            &Cruise LLC'22 \cite{cruise2022safety} & LiDAR, radar, camera, etc. & HD map & map-based, multi-modal \\
            &Nvidia'24 \cite{nvidia2024safety} & LiDAR, radar, camera, etc. & HD map & map-based, multi-modal \\
            & Mobileye'24 \cite{mobileye2024safety} & LiDAR, radar, camera, etc. & HD map & map-based, multi-modal \\
            & Tesla'21 \cite{Tesla2021} & Cameras & \xmark & mapless, vision-based  \\
            & Toyota'22 \cite{Toyota2022} & LiDAR, radar, camera, etc. & HD map &  map-based, multi-modal \\
            & Mercedes-Benz'23 \cite{Mercedes2023} & LiDAR, radar, camera, etc. & HD map &  map-based, multi-modal \\
            & Wayve'25 \cite{hawke2021reimagining, WayveAV2} & camera, radar, (opt. LiDAR)  & \xmark & mapless, vision-first \\
            \midrule
            \multirow{4}{0cm}{\rotatebox{90}{V2X}}& Could-CEM'20 \cite{lampe2020reducing} & internal multi-modal & multiple multi-modal vehicles& collective env. model\\     
            &UULM-Bosch'21 \cite{buchholz2021handling} & LiDAR, radar, camera & HD map, infrastructure sensors & multi-modal, connected \\
            & Providentia++'23  \cite{cress2023intelligent} & onboard sensors & multi-modal infrastructure RSU & real-time digital twin\\
            & V2X-Percep.'24\cite{shan2024experimental, lampe2020reducing} &onboard sensors & infrastructure sensors & multi-modal, multi-system  \\
			\bottomrule \\[-6pt]
	\end{tabular}}
\end{table*}

In  Table \ref{tab:comparison_AI_method_inputs} it becomes visible that the investigated methods do not match the real commercial system configurations. While in real systems, except Tesla \cite{Tesla2021} and partly Wayve \cite{hawke2021reimagining, WayveAV2}, a wide range of internal sensor inputs are incorporated \cite{uber2020safetyreport, lyft2020safety, ArgiAI2021safety, Motional2021safety, waymo2021safetyreport, zoox2021safety, cruise2022safety, nvidia2024safety, mobileye2024safety, Toyota2022, Mercedes2023}, the investigated methods mostly focus on vision- or LiDAR-based systems. Even in multi-modal systems, LiDAR and camera are only incorporated  \cite{jia2023think, pomerleau1988alvinn, chitta2022transfuser, jia2023driveadapter, chen2022learning, xu2024drivegpt4, you2024v2x, ye2023fusionad}. Although V2X research is expanding information acquisition, it is neither part of the AD stacks currently under research nor part of the real systems.

However, increased interconnectivity facilitates the versatile use of external data and information. Beyond the information currently considered, such as to extend the vehicle’s FOV, high-definition (HD) maps \cite{wong2020mapping, elghazaly2023high} offer further possibilities by providing rich situation semantics. HD maps can serve as virtual sensors, aggregating knowledge from physical sensors and prior data to represent the environment. They contain information about, for example, an adjacent building being a school, about a bus stop across the street, or about being on a highway. Furthermore, besides speed limits also information regarding traffic-calmed areas or permanent traffic barriers can be stored in HD map \cite{elghazaly2023high}, supporting perception, planning, and decision-making in automated vehicles. Beyond that, there is the possibility that through increased interconnectivity and cooperation additional information such as the global destination of other local participants is known. All this information represents extensive context knowledge, which can be taken into account. Such external sourcing of context offers the opportunity to reach enhanced situation understanding, required for nuanced SA and decision-making for the desired level of autonomy. However, it also increases the burden of targeted data processing and information aggregation while simultaneously increasing dependencies. Although this poses challenges, the external sourcing of context offers the opportunity to minimize internal expenses, e.g. the recognition of a school. One approach that address the respective challenges is the continued consideration of external sources within the context layer discussed above. Nevertheless, a measure of trustworthiness should be included for external data in accordance to a respective security level in order to ensure reliability and integrity. Especially the mutual reassurance between external information and internal awareness represent a hedging methodology, with a variety of consecutive open research questions.

\subsection{Hypothetical Structure of Architecture} 
\begin{figure*}[]
	\centering	
	\includegraphics[width=\linewidth]{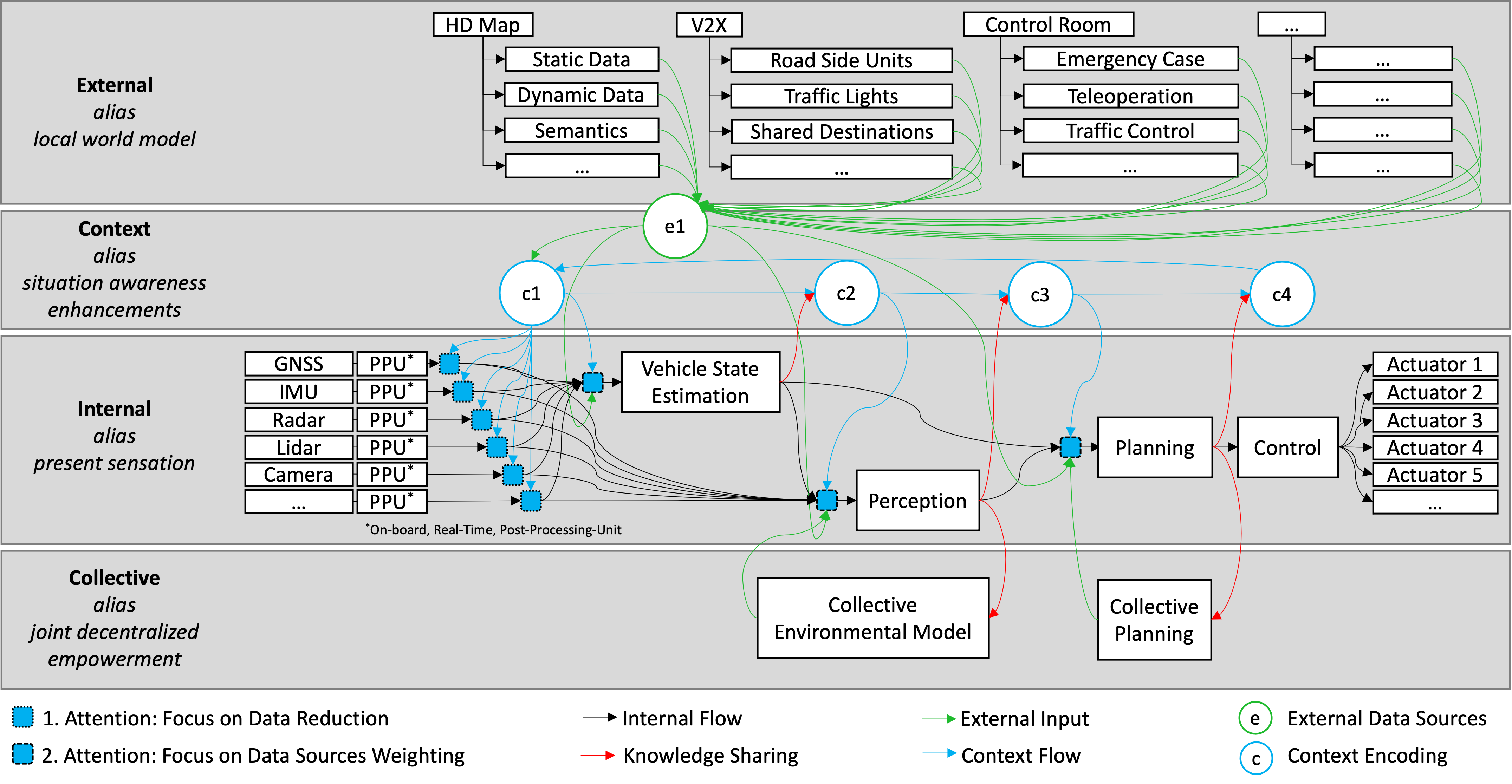}
	\caption{Possible architecture integrating situational awareness.}
	\label{fig:hypo}
\end{figure*}

A conceivable consideration of the previously outlined aspects is sketched in Figure \ref{fig:hypo} in an abstract hypothetical sense. Similar to a meta-learning approach like conditional neural processes \cite{garnelo2018conditional}, the external knowledge at hand (HD map, V2X, control room, ...), referred to as the local world model, could be summarized (e1) and encoded into a context state (c1). This context state could be conglomarate of various representations like VLM and LLM driven tokens, but also latent or semantic representations. Building on such a context (c1), it could be used in the first attention stage to context-dependently constrain the information density of the pre-processed raw data. In the second attention stage, the weighting of the individual sensor modalities could be adjusted depending on the context. For example, when turning right at an intersection without any other participants, the first stage could specifically adjust the perception task in a context-specific manner, while realizing a data reduction. The second stage, on the other hand, could take, e.g.,  weather conditions into account and reduce the impact of sensors that exhibit weather-related degradation via adjusted weights. 

In addition, the newly acquired context could be fed to the context state by internal processing steps in order to enrich it. Besides that, it is also imaginable that collective environment models and collaborative planning are directly taken into account. Also, general external information (e1) can be used as input for the respective processing sub-modules, whereby the context is a crucial factor in pre-filtering their significance and exploitation. Overall, the context is decisive with respect to what extent the respective data, information, or even knowledge is to be considered. 

Note, Figure \ref{fig:hypo} illustrates a unique conceptual framework of a hypothetical approach that takes into account the flexibilization of modules through attention driven query-based interfaces, the increase of SA through a consistent and continuously updated context representation, the direct and indirect consideration of external sourcing, and thus takes up various previous aspects. However, the illustrated architecture does not represent a solution but rather serves as a linked representation of the aforementioned aspects for further discussion. 

\subsection{Perspectives \& Implications on AD Stack Developments} 

The aforementioned conglomerate of observations shows that an increase in autonomy is accompanied by an increase in system complexity, which in turn requires a change in the AD stack. Initially opposing approaches such as modular service-oriented AD stacks and vanilla E2E AD stacks are converging more and more via I-E2E and M-E2E-P and combining the respective advantages. However, internal and external sourcing as well as V2X communication and collaboration are limited. In this area of tension, query-based interfaces can be further developed from simple interfaces to more comprehensive interfaces and even orchestrators that control the flow of information. In this way, different sources and functional capabilities can be queried and taken into account via query-based attention mechanisms. Increasing coupling with FM, for example, would also be conceivable. If internal uncertainty increases significantly, for example, the LLM and its reasoning capabilities might be called upon. 

One major implication with respect to corresponding further developments is the increasing need to train learnable interfaces and orchestrators. Here, a combination of bottom-up modular functionalities that allow for permissible functional and expert knowledge without affecting the generalization capability of the overall system, together with a learning-oriented top-down approach that addresses the inherent complexity by means of an implicit data-driven approach, seems promising for closing the semantic gap and approaching towards the desired autonomy. In addition, there is the possibility that the efforts of modular service-oriented architecture, such as the use of quality vectors \cite{ecksteinautotech}, that specifically supplement payload data, in order to increase reliability, could be integrated as a self-assessment. For instance, self-assessment quality data could be integrated across the stack while simultaneously guaranteeing interpretability along the module interfaces or branched out semantic representations. We refer to such an increasingly self-responsible yet interpretable and debuggable architecture, which represents a combination of modular service-oriented and (M-)E2E architecture, as a service-oriented modular end-to-end (SO-M-E2E) architecture.

\subsection{Implications on  Safeguarding} 
 
In the context of interpretable AD stacks, the bottom-up approach is primarily pursued, integrating heavily extracted system and expert knowledge into the architecture and tech stack \cite{ecksteinautotech}. However, as discussed fully autonomous vehicles require new degrees of freedom. The consequences of underlying paradigm shifts, such as the transition from hard-coded connections to flexible soft-coded connections, illustrate changed framework conditions which, in addition to development, also have an impact on safeguarding. 
 
Modular assurance, e.g. \cite{stolte2020towards, klamann2023introducing}, is a target-oriented approach in the context of a modular service-oriented architecture in AD. Moreover, recent approaches acknowledge the increasing use of learned components by means of data-based interface definitions for modular safety approval \cite{blodel2024towards}. However, the current modular assurance methodology is expected to reach limits in fully autonomous driving with respect to accounting for circular reasoning. Nevertheless, the modular assurance and in particular the data-driven approach of \cite{blodel2024towards} provides a basis for extensions that could address circular reasoning on this basis. Since prospective architectures for fully autonomous driving, such as the hypothetical architecture (Figure \ref{fig:hypo}), or the discussed SO-M-E2E architecture in general, must necessarily take into account the given context. Here as well, the targeted supplementation of existing bottom-up approaches with deliberate top-down approaches to address the emerging challenges is conceivable. In this regard, RL, CIL and teacher-student approaches of top-down vanilla E2E approaches could be integrated on or along the SO-M-E2E architectures in the future for this purpose.

\subsection{Discussion}
In line with \cite{chen2023end}, it could be assumed that future highly autonomous systems will need generalizing AI components, especially to address the multitude of corner cases at a higher level. However, this does not mean that task-specific, narrow AI systems or working engineered subsystems should be eliminated. Instead, these sub-module-specific high-performing systems should be retained, which also facilitates better explainability and debugging. However, generalizing AI offers the advantage of being able to respond autonomously to changing environments as well as to changing availabilities of external information. In addition, multiple perceptual modalities could be addressed across different vehicles. In the course of generalizing AI, approaches such as few-shot \cite{wang2020generalizing, kadam2020review}, meta-learning \cite{hochreiter2001learning, finn2017model, vanschoren2019meta}, and FMs are currently very popular. Combined with tightly connected AI systems and extensive knowledge-informing data and expertise that can be integrated top-down, highly AD could achieve astounding performance. In this way, today's obvious problems of automated vehicles in the United States \cite{templeton2023cruise, californiaProbe2023, NHTSARecall23E} could be addressed, thus achieving the desired resilience and robustness. In this regard, an SO-M-E2E architecture seams to be promising concept for ongoing research. Moreover, it also may be noted that the SO-M-E2E architecture could be interpreted as the first concretization of Yann LeCun's world model of a human-like AI \cite{lecun2022path}, but tailored towards autonomous driving. To be specific, these are recurrent, object-driven world models, that to a certain extent, mimic model predictive control (MPC) in neural networks. In this way, intelligence in the manner of Konrad Lorenz \cite{lorenz1973ruckseite}, thus operating in imaginary space, could be realized. 

Furthermore, the increasing use of AI across modular system boundaries, shows the increasing importance of corresponding data, both for system development and for verification and validation. This was analyzed and discussed in the context of \cite{ullrich2024expanding}, whereby a corresponding concept for an iterative development, verification, and validation of emerging complex systems that include AI was presented with specific emphasis on data relevance. Moreover, in this regard, appropriate databases that could be used exist for many years, especially for accident data. For example, the Fatal Analysis Reporting System (FARS), the Crash Report Sampling Systems (CRSS) or the Crash Investigation Sampling Systems (CISS) and others as shown in \cite{kusano2023comparison}. Furthermore, manufacturers of automated vehicles are required to report crash data. Accordingly, a transition within the field of safety analysis towards data-based AI safety assurance \cite{ullrich2024aisafety_assurance} is also foreseeable. 
\section{Prospect Challenges and Opportunities}\label{FutureAD}
The previous sections briefly outlined the current status of AD, highlighted current limitations, and presented prospects for advancing towards fully autonomous driving. Up to this point, the focus has primarily been on functionality and technological possibilities. However, the widespread deployment of autonomous driving necessitates scalability across various vehicle setups and execution environments, demanding flexibility for efficient adaptability. Therefore, this section extends the discussion to application-specific hurdles that have a decisive impact on the widespread availability of autonomous driving, identifying both prospect challenges and opportunities.

\subsection{Challenges of Prospective Autonomous Driving}

As Section \ref{AutonomousDriving} showed, there are various aspects that would enable the limitations to be overcome and autonomy to be increased. Regardless of the specific technological realization, it is becoming visible that the use of AI and the importance of data appear to be increasing. This results in specific challenges for the present application, which are examined in more detail below.

Among other things, compliance with the applicable rules and laws is crucial for the introduction of autonomous driving. Besides compliance with existing regulations, the expanding regulation of AI poses a considerable challenge. This is especially true as different AI regulatory approaches are being pursued \cite{metiaigovernanceguidelines, dsiaiframework, eu_parliament_2024corr, billc27, aigovernanceprinciples, ai2023artificial} and the regulatory landscape tends to remain dynamic and heterogeneous in the near future. This leads to the subsequent challenge of ensuring autonomous driving takes various regulatory requirements into account, e.g. by means of a system design integrated adaptive verification procedure of obligations.

In order to outline some regulatory requirements, the following mentions a few aspects of the recently enacted EU AI Act \cite{eu_parliament_2024corr}, which is one of the first far-reaching laws on the use of AI. This law takes a risk-based approach, whereby autonomous driving is considered a high-risk system and is therefore subject to numerous obligations (e.g. Articles 16, 18, 23, 24, 25, 26). Worth mentioning here is the required risk management system (Article 9), which should consist of a continuous, iterative process throughout the entire lifecycle. Also worth mentioning are the requirements for data and data governance (Article 10), which set demands for training, testing and validation datasets, quality criteria, design decisions and data collection processes. In addition to these aspects, many others are subject to requirements such as human oversight (Article 14), post-market surveillance (Article 72), and serious incidents reporting (Article 73). The above illustrates the comprehensive regulatory demands based solely on the EU AI Act. It is to be expected that the regulatory landscape will (initially) be diverse and changeable across the globe \cite{ullrich2024aisafety_assurance}.    

\textit{\textbf{Derived need:} Autonomous driving should already take different regulatory requirements into account at the design stage and offer flexibility with regard to the verification of regulatory requirements in order to be prepared for different markets and changes over time.}

Another challenge, which is related to both the regulatory requirements and the technical system, is safeguarding. Especially if AI is used along the critical path from perception over decision-making to execution, which is likely according to previous analyses. In this context, established standards such as ISO 26262 (Functional Safety, FuSa) \cite{iso26262}, ISO 21448 (Safety of the Intended Functionality, SOTIF) \cite{iso21448}, ISO/PAS 8800 (Safety and AI in road vehicles) \cite{iso8800}, and UL 4600 (Safety of Autonomous Products) \cite{isoUL4600} provide foundational guidance but remain difficult to apply consistently to complex, data-driven AI systems. Consequently, new concepts such as the AI Safety Integrity Level (AI-SIL) \cite{diemert2023safety} have emerged, highlighting the need for a shift towards data-driven AI safety assurance \cite{ullrich2024aisafety_assurance}. In this regard, a complementary perspective based on control-theoretic methodologies has recently been proposed \cite{ullrich2025new}. Even though this approach recognizes the fundamental shift from explicit mathematical to implicit data-driven systems, along addressing the broader challenge of predominantly non-agnostic safety assurance methods \cite{neto2022safety}, many open research questions regarding AI safety remain \cite{amodei2016concrete}.  

Besides the general challenges, there are also application-specific challenges in using AI for autonomous driving. In particular, due to the data-driven modeling foundations and assumptions, continuously checking the completeness of datasets is essential for applications operating in open, long-tail distributions, such as AD. Companies such as Waymo and Tesla address these challenges through iterative, lifecycle-oriented processes aimed at continuously improving system performance. While Waymo’s Safety Determination Lifecycle \cite{favaro2023building} provides a more generic framework, Tesla’s Data Engine \cite{karpathy_cvpr21} specifically targets the particularities of AI systems. To integrate these frameworks, an extension of the V-model for iterative, data-based development has been proposed, representing a generalized process reference model for complex systems incorporating AI \cite{ullrich2024expanding}. These approaches can be regarded as risk management systems under Article 9 of the EU AI Act, highlighting the importance of iterative improvement alongside addressing AI safety concerns. As noted in \cite{kusano2023comparison}, such processes support both prospective safety analysis and retrospective evaluation.

\textit{\textbf{Derived need:} Autonomous driving should be designed such that iterative further developments and releases are possible in an efficient manner throughout the entire lifecycle.}

In line with the requirement for an iterative refinement process, data is an crucial basis for developing and ensuring the safety of complex, AI-inclusive systems such as autonomous driving. Thereby, data from critical situations is of particular interest, which is why data related challenges should be taken into account at an early stage.

Considering automated and autonomous vehicles operating around the globe, harvesting and sharing data, there is a myriad of data available. This data includes data from different countries with different traffic rules and behaviors. Furthermore, the data originates from different vehicle setups, types, and manufacturers, which is why different sensor modalities, types, and qualities serve as sources, and a wide variety of data is available. Thus, according to such a data harvesting convention, data of different characteristics from relevant and critical situations is available. 

\textit{\textbf{Derived need:} The various trigger data should be generically usable, but also transferable to different use cases, e.g. vehicle configurations and environments.}

While the previous aspect refers to the diversity of systems and environments in terms of data, it can also pose challenges in terms of functional transferability. In this respect, human-like autonomy is desirable as humans are able to adapt their abilities, e.g., perception, decision-making, and planning, to the given circumstances. For example, when humans switch vehicles, they can adapt their perception and planning. Even if they change the execution environment, e.g., to a different country with different rules and behavior patterns, they can adapt and operate safely in this different open world. 

\textit{\textbf{Derived need:} Human-like autonomy of functionalities, especially from perception to trajectory planning, is desirable. The aspired autonomy should be adaptable to different Operational Design Domains (ODDs), i.e., domains that define the specific conditions under which an AD system can operate safely. Furthermore, the autonomy should also be able to handle vehicles with different equipment, such as sensors and actuators, to enable efficient and scalable development and deployment of autonomous driving.}


In addition to the technological and regulatory challenges, there are also socio-political challenges. These include, for example, the socio-political definition of acceptable risks, but also society's trust in autonomous vehicles. Therefore, participation in the discourse on ethical issues and the agreement on acceptable risks demand a comprehensible, understandable, and publicly transparent development and validation process. Furthermore, traceability and understandability, also referred to as human-interpretable, are the basis for enabling efficient assurance and approval procedures.

\textit{\textbf{Derived need:} Comprehensible and understandable processes as well as a publicly accessible safety argumentation are aspirational for socio-political questions of trustworthiness and thus the acceptance of autonomous vehicles. Likewise, the behavior of the vehicles must be interpretable, comprehensible and acceptable, not only for passengers but also for external parties.}

\subsection{Countermeasures and Opportunities for Effective Transferability and Efficient Scalability}

While the previous subsection outlines challenges of prospect autonomous driving and derived needs, this subsection outlines opportunities alongside the consideration of the previous derived needs. 

According to the desirable human-like autonomy, it is preferable to consider the core functionalities for safe and admissible trajectory planning independent of countries and data sources. This idea could be addressed by training a baseline perception-to-trajectory model (P2T), e.g., on the basis of a SO-M-E2E stack, which is primarily designed to generate trajectories in a variety of situations. In this context, a P2T model could be trained similar to a language model, which first learns the language and then in a further step post-trains the meaning, correctness, and ethics. Nevertheless, classical language and foundation models are not capable of human-like internal planning. In contrast, the hypothetical architecture from Section \ref{AutonomousDriving}, or more generally the SO-M-E2E architecture, takes existing submodules, adapts their use in the overall AD stack, and represents a first approach for a foreseeable applicable realization of objective-driven recurrent world model that offer human-like planning \cite{lecun2022path}. Consequently, in the course of the development of such a P2T system, the cognitive interfaces must initially be trained on the basis of the general data. This P2T world-model, as we call it, thus draws on a wide range of knowledge. However, such a model cannot correctly meet the requirements of a dedicated country-specific ODD. This is underscored by findings from a transfer learning study on motion prediction \cite{ullrich2024transfer}, which suggests that multi-task learning performs less effectively across diverse contexts compared to targeted fine-tuning tailored to respective settings. Therefore, as with the human, adaptation to the ODD at hand is necessary. Furthermore, an adjustment to the system, the vehicle at hand, is also inevitable.

The adaptation to the ODD at hand requires data of the specific ODD. At the same time, the worldwide found trigger conditions from different ODDs should be taken up and considered. In this context, the current state of the art is a scenario-based approach that extracts, defines, and generates corresponding scenarios, often manually. To minimize human hand engineering and the corresponding human-based impact, an automated procedure would be desirable. In this course, the first goal is to transfer all relevant and critical cases of each ODD into the ODD at hand. The second goal is to provide a unified all-modality perception. In other words, every relevant type of data source considered in autonomous vehicles should be available. 

Considering, for example, a critical situation detected in the USA by a Tesla, the corresponding recorded data would not contain radar nor lidar data, as Tesla pursues a vision only approach. However, the ODD of the system under development could indicate Germany, among others. Accordingly, the first goal is to transfer the critical situation to a German ODD. Therefore, the scenario is to be generated with German traffic signs and road markings. The second goal concerns the sensor modality, here the objective is that all modalities are available to use the data across divers vehicle setups. Consequently, synthetic radar and lidar data is also required, as the Tesla recorded dataset does not contain those. In order to avoid recreating each situation individually in a simulation environment, an automated procedure is proposed. This should generate synthetic scenarios that ultimately represent an artificial but realistic synthetic simulation. In the best case, a system is available that transfers all global trigger conditions to the ODD under development and artificially generates missing sensor signals. The result would be an ODD-specific master dataset that maximizes the provision of information. In this way, the challenges of different application areas and different legal and law enforcement areas could be taken into account. For instance, the realization of such a scenario generator is closely related to SE. Within the field of SE possibilities of advanced AI methodologies and systems, such as OpenAI's Sora, are discussed in terms of imaginative intelligence \cite{wang2024does}. In particular, the emerging imaginative intelligence offers great potential towards a scenario generator, and in general, a even more data-driven development and verification and validation in alignment to \cite{ullrich2024expanding}. 

Thereby, implicit considerations of expert knowledge about safe and ethical behavior in the respective operational area, e.g., Germany, would be worthwhile. Nevertheless, while transferring corresponding ODD-specific reasoning and decision-making to the system, it would be beneficial to eliminate the need for human explanation of relationships. The advantage of that would be, that rules do not have to be explicitly extracted and communicated to the system. Especially in corner cases, there is usually an overlay of rules where humans are able to make instant decisions based on situation understanding \cite{brehmer1992dynamic, canas2003cognitive}. However, even humans have difficulty describing the correct behavior in tough situations based on rules. Moreover, there are usually several different rule reasonings that lead to very similar system behavior. Rather than extracting all the rules in each situation, prioritizing them, and possibly coming to a halt due to conflicting rules, the goal of the autonomous system is to find a safe and appropriate solution in each situation. Meaning, to realize safe instant decision making based on enhanced contextual awareness. Therefore, new approaches are needed for the integration and learning of high-level reasoning and decision making. In this context, broad accessibility and thus the social discussion of a permissible acceptable risks is also of significance. Overall, a procedure that simplifies the approval process while considering regulatory requirements, such as human oversight of EU AIA, along increasing social acceptance is worth striving for. 
 
While the process outlined so far illustrates how a model can be developed that operates according to specific traffic signs, traffic rules, behavioral rules, as well as the regulations of a particular ODD, the model still assumes all sensor modalities to be given. In a next step, the model that has achieved simulative safety on synthetic data can be adapted to a concrete system. Data from the system's sensors can be used in this process. Since each manufacturer uses different sensor setups, there will likely be an adjustment in sensor modalities and data characteristics. During this model transfer, the model can be fine-tuned and optimized to work with the given sensor setup. Thereby, the methodology provides the ability to evaluate relevance and sensitivity for individual sensor modalities. Thus, minimum requirements for the sensors are implicitly stipulated. After certification of transfer reliability and implicit release of the sensor setup, evaluation and release can take place in the real vehicle. Thus, the methodology offers the possibility to adapt an ODD-specific P2T model to a variety of vehicles of different vehicle bodies, types and manufacturers.

Ultimately, it is conceivable to integrate the individual process steps into an iterative cycle, as outlined within \cite{ullrich2024expanding}, that enables continuous refinement and adaptation to the changing real world. Moreover, it is demonstrated that data is a central component. Furthermore, the prospect enables resources to be pooled centrally and collectively drive the core of autonomy as well as define and ensure safety across the board. In terms of model transfer, each vendor can in turn access and fine-tune the system. In terms of governance, this creates incentives for data sharing, although critical data sharing is considered to be essential and mandatory according to divers regulations. Moreover, it is also imaginable to establish a federated learning concept in order to protect vendor- and customer-specifics despite sharing data in general.

Centralization and concentration reduce technical effort. By pooling data, a better system might be created than any one individual can create, at a lower cost and in less time. Licensing fees can offset the cost of creating country-specific models while ensuring safety across all manufacturers. The world model and country-specific models could be developed and overseen by an international, interdisciplinary group. Thus, in the spirit of open innovation and close collaboration, companies, countries, and organizations could work together towards safer mobility.

In addition to the countermeasures and opportunities mentioned so far, a separate option is discussed below in order to do justice to the diversity of possibilities. These are large foundation models, which dispense with dedicated sub-modules by solving the entire task in an end-to-end manner. This seems to be radical at the moment, but in the past, recurrent models such as RNN, GRU, LSTM and also transformer models, which were initially introduced via natural language processing in the field of AI, have become very important in technical modeling. In line with this are the latest developments such as Nvidia's Alpamayo-R1 \cite{wang2025alpamayo}. Ultimately, these approaches can also only be realized with a corresponding amount of data and computing power. In particular, the combination with few-shot  learning techniques promises great potential, although these models are currently still difficult to interpret and construct, which is reflected, for example, in the large variance of the parameter sizes used. However, these methods are a possibility. And indeed, we can already see some first realizations of trajectory predictions \cite{seff2023motionlm} and autonomous driving in general \cite{cui2023drivellm}. In addition, LLM can also contribute to the human interface in the future of autonomous driving \cite{huang2023visual}. Although this seems promising, we share the view of \cite{lecun2022path}, stating that a human-like AI requires a higher level of intelligence and must be able to plan and reasoning internally.

\section{Discussion}\label{ADDiscussion} 

While there is a large number of review articles in the field of AD \cite{yurtsever2020survey, ma2020artificial, grigorescu2020survey, huang2020survey}, the rapid technological advancement requires a continuous review of the field. In this regard, the significant development of AI methods \cite{wang2020generalizing, batarseh2021survey, hospedales2021meta, zhou2022domain} are leading to breakthroughs in the field of AD. Accordingly, existing approaches are constantly being reassessed. Recently, FMs such as LLMs, VLMs, or VLAs have raised new possibilities and questions. First reviews already deal with these questions \cite{zhou2024vision, yang2023llm4drive, kim2024openvla, wang2025alpamayo}. At the same time, it is becoming apparent that the difference between the AD systems currently being researched and the more classic, modular, service-oriented AD systems is widening. Research is also very fragmented. Accordingly, a large number of reviews look at individual sub-research areas such as functional subareas \cite{paden2016survey, jin2019advanced, gulzar2021survey, gupta2021deep, huang2022survey, hagedorn2024integration}, V2X connectivity \cite{wang2018networking, alnasser2019cyber} and collaboration \cite{datondji2016survey, cress2023intelligent}, and more \cite{ignatious2023analyzing, ding2023survey, ullrich2024aisafety_assurance}. In addition, most of the studies are in the predefined SAE automation stages.

This review therefore differs from existing reviews in several dimensions. Firstly, it builds a bridge between service-oriented AD stack systems and the latest developments. Building on this, it anticipates opportunities and the resulting challenges. While the focus is on the functionalities and implications of AI, it also builds a bridge to broader aspects such as transferability, scalability as well as socio-political aspects. Finally, this review also distinguishes itself through its in-depth SA and cognition-guided analysis of existing capabilities and necessary developments. This functional and cognitive perspective facilitates decoupling the underlying system requirements and systematic challenges from technical solutions to maintain validity for further developments. At the same time, the current technological possibilities are also taken into account. This functional and holistic approach across several dimensions distinguishes the present study from existing ones and thus offers a new perspective on the field.

\section{Conclusion}\label{ADConclusion}

It was shown that the modular service-oriented software architecture is increasingly complemented by AI methods and thus performance can be improved, but certain structural limitations are still integral. The analysis highlighted that increased autonomy requires particularly to address the challenge of the open long-tail distribution of corner cases in the real world and therefore the overcoming of rule-based approaches and system orchestration. Furthermore, this revealed that SA is of particular importance in order to imitate humans in terms of dynamic, adaptive, and context-specific decision-making. At the same time, the analysis resulted in relevant requirements that are necessary for corresponding improvements. 

A possible architecture that takes SA integral into account was conceptualized based on the findings. This concept considers attention-based mechanisms as a central enabler that ensures the desired flexibility, adaptivity, and context specificity at the orchestration level without relying on human-extracted and hand-crafted rules. This is promising for achieving the desired human-like responsiveness in corner cases that are previously unknown. At the same time, the possible architecture raises open research questions regarding its realization. Primarily with regard to the generation, aggregation, and use of related data for the training of such a cognitive orchestrator. Possibilities such as federated learning or end-to-end learning have also been identified in this regard. 

In relation to the existing AD stack, the proposed SO-M-E2E architecture can be interpreted as a coherent extension that integrates existing knowledge and functionalities. Also with regard to the quality vectors proposed in \cite{ecksteinautotech}, it can be considered as a possible extension, as the cognitive orchestration ultimately resembles quality vectors in an integral, human-like manner. In addition, further challenges and opportunities from regulation and socio-political issues to commercial aspects of scalability and transferability were examined and open research questions outlined. 

In general, it can be seen that AI offers many opportunities at a functional perspective, but also raises corresponding research questions. Overall, however, it is clear that AI and the use of AI is a promising tool for achieving previously unattainable levels of autonomy, albeit with implications for the AD stack as well as respective safeguarding, transferability, and scalability. Accordingly, a large number of open research questions have been identified, which need to be addressed in the future while promising significant added value. 

\bibliographystyle{IEEEtran}
\bibliography{literature}

\vspace{- 10mm}
\begin{IEEEbiography}[{\includegraphics[width=1in,height=1.25in,clip,keepaspectratio]{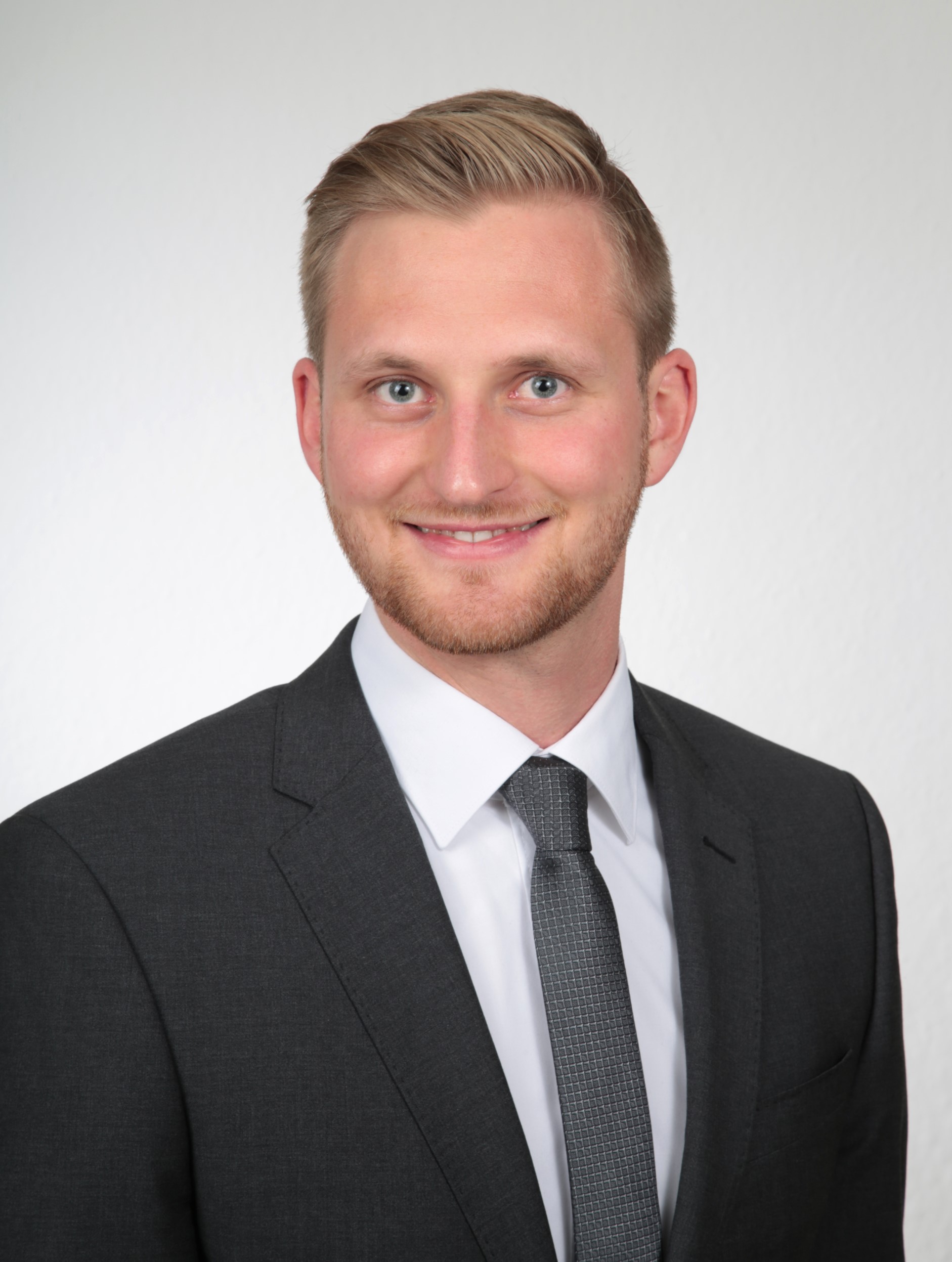}}]{Lars Ullrich}
	received the M.Sc. degree in mechatronics and the Ph.D. (Dr.-Ing.) degree from Friedrich–Alexander–Universit\"at Erlangen–N\"urnberg, Germany, in 2022 and 2025, respectively. His research has focused on probabilistic trajectory planning for safe and reliable autonomous driving in uncertain, dynamic environments, with a particular emphasis on challenges arising from the use of AI in automated driving. His current research interests center on safe embodied AI. Since early 2025, he has been elected Vice-Chair of the IEEE Intelligent Transportation Systems Society Germany Chapter.
\end{IEEEbiography} \vspace{-10 mm}
\begin{IEEEbiography}[{\includegraphics[width=1in,height=1.25in,clip,keepaspectratio]{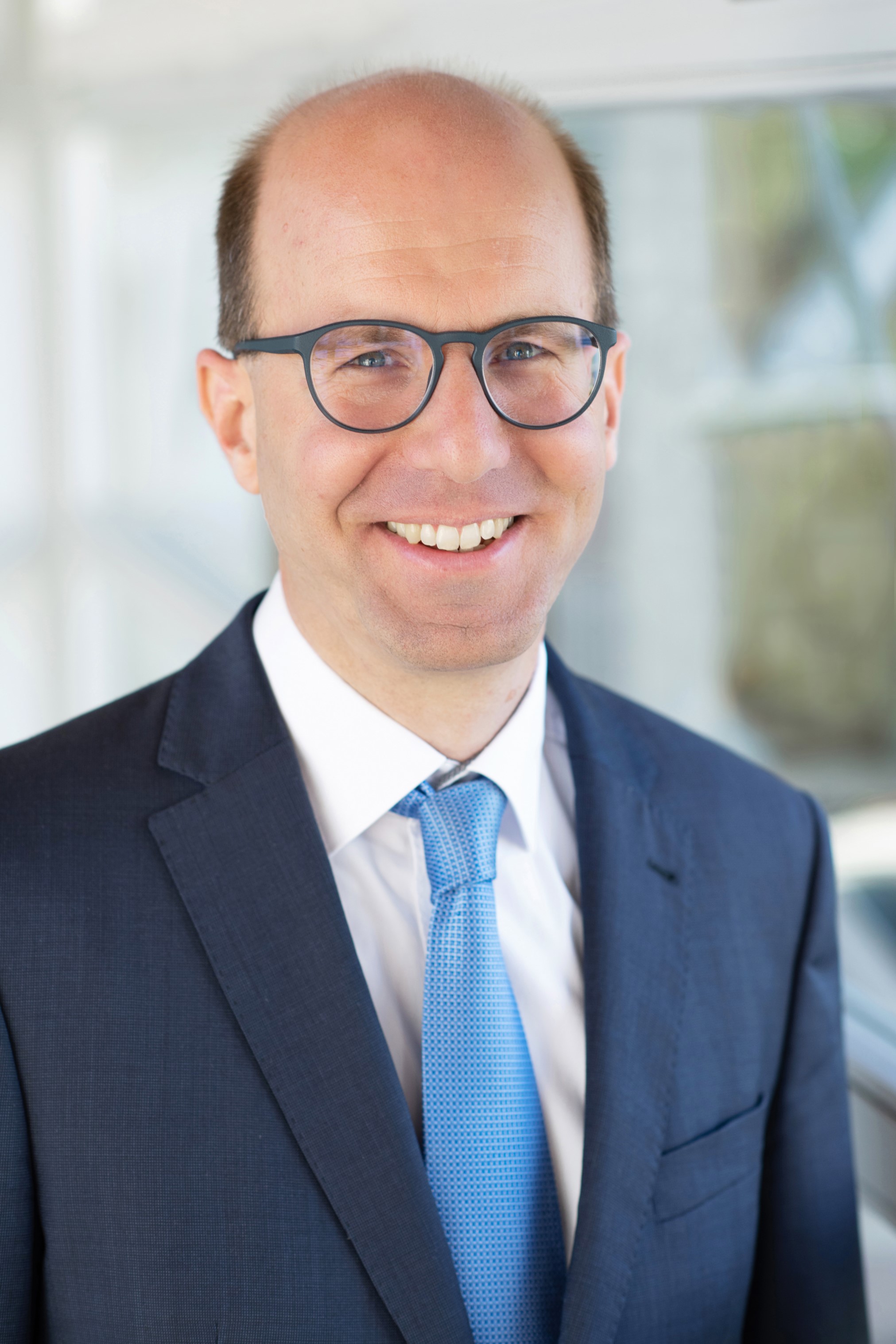}}]{Michael Buchholz} 
	received his Diploma degree in Electrical Engineering and Information Technology as well as his Ph.D. from the Faculty of Electrical Engineering and Information Technology at University of Karlsruhe (TH)/Karlsruhe Institute of Technology, Germany.  He is a research group leader and lecturer at the Institute of Measurement, Control, and Microtechnology at Ulm University, where he earned his \qemph{Habilitation} (post-doctoral lecturing qualification) for Automation Technology in 2022 and the title of an \qemph{apl. Professor} (adjunct professor) in 2025. His research interests comprise connected automated driving, electric mobility, modelling and control of mechatronic systems, and system identification.
\end{IEEEbiography} \newpage
\begin{IEEEbiography}[{\includegraphics[width=1in,height=1.25in,clip,keepaspectratio]{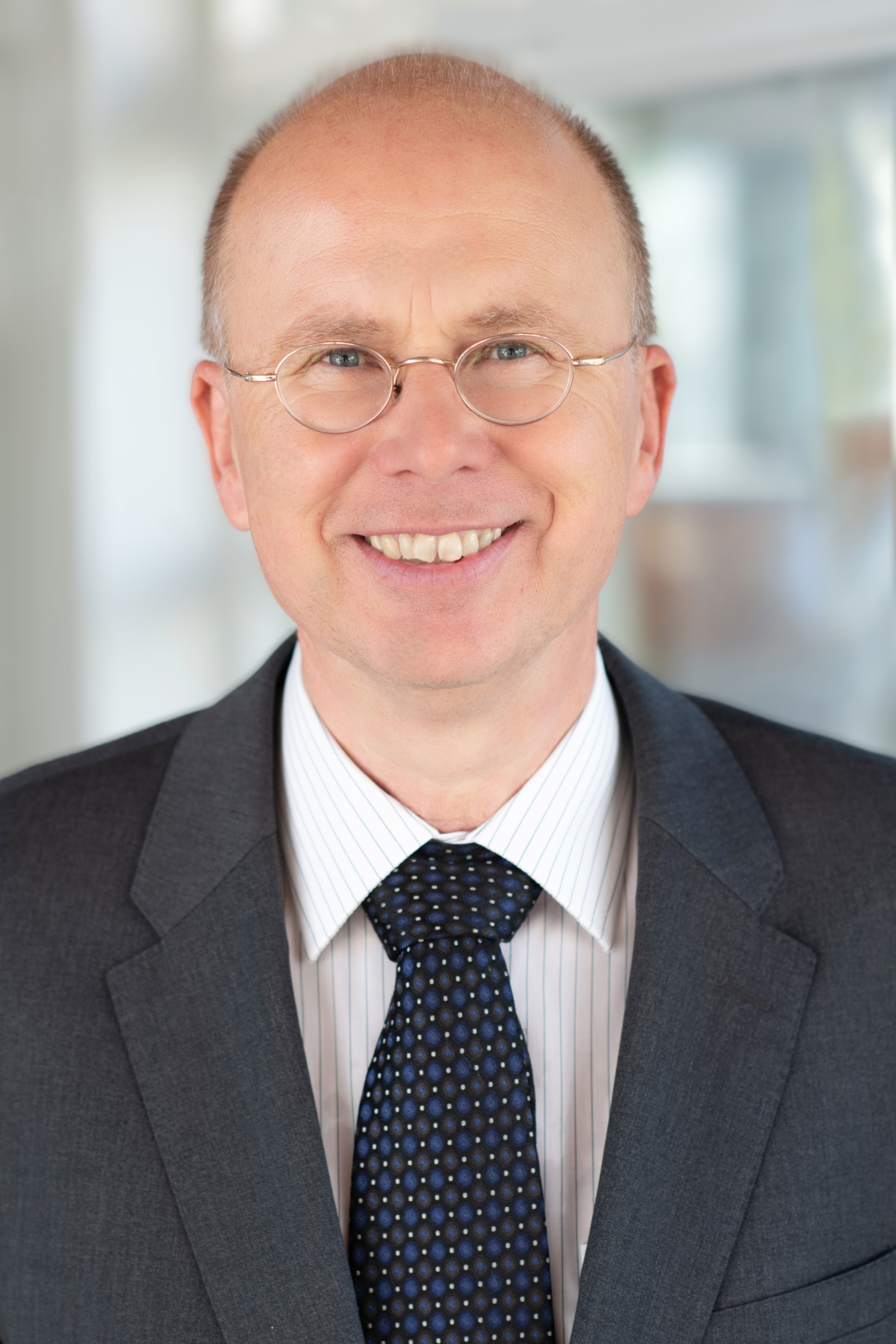}}]{Klaus Dietmayer}
	(Senior Member, IEEE) earned his degree in electrical engineering from the Technical University of Braunschweig, Germany and completed his Ph.D. in 1994 at the University of the Armed Forces, Hamburg, Germany. Afterwards, he began his industrial career as a research engineer at Philips Semiconductors, Hamburg, progressing through various roles to become the manager for sensors and actuators in the automotive electronics division. 
	
	In 2000, Dietmayer was appointed as a Professor of Measurement and Control at the University of Ulm. He currently serves as the Director of the Institute for Measurement, Control, and Microtechnology within the School of Engineering and Computer Science.
	
	His primary research interests include information fusion, multi-object tracking, environment perception, situation assessment, and behavior planning for autonomous driving. The institute operates three automated test vehicles with special licenses for public road traffic, along with a test intersection equipped with infrastructure sensors for evaluating automated and networked cooperative driving in Ulm.\\
\end{IEEEbiography} \vspace{-85 mm}
\begin{IEEEbiography}[{\includegraphics[width=1in,height=1.25in,clip,keepaspectratio]{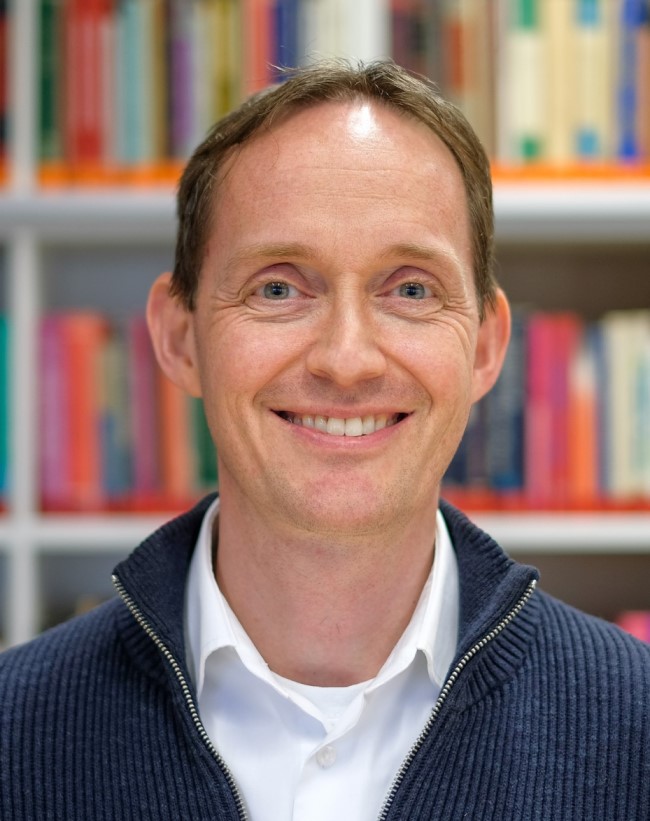}}]{Knut Graichen}
	(Senior Member, IEEE) received the Diploma-Ing. degree in engineering cybernetics and the Ph.D. (Dr.-Ing.) degree from the University of Stuttgart, Stuttgart, Germany, in 2002 and 2006, respectively.
	
	In 2007, he was a Post-Doctoral Researcher with the Center Automatique et Syst\`emes, MINES ParisTech, France. In 2008, he joined the Automation and Control Institute, Vienna University of Technology, Vienna, Austria, as a Senior Researcher. In 2010, he became a Professor with the Institute of Measurement, Control and Microtechnology, Ulm University, Ulm, Germany. Since 2019, he has been the Head of the Chair of Automatic Control, Friedrich–Alexander–Universit\"at Erlangen–N\"urnberg, Germany. His current research interests include distributed and learning control and model predictive control of dynamical systems for automotive, mechatronic, and robotic applications.
	
	Dr. Graichen is the Editor-in-Chief of Control Engineering Practice.
\end{IEEEbiography}

\EOD

\end{document}